\PassOptionsToPackage{sort&compress}{natbib}
\documentclass{article}

\usepackage{arxiv}

\usepackage[utf8]{inputenc}
\usepackage[T1]{fontenc}
\usepackage{hyperref}
\usepackage{url}
\usepackage{booktabs}
\usepackage{longtable}
\usepackage{threeparttable}
\usepackage{float}
\usepackage{amsmath,amssymb,amsfonts}
\usepackage{nicefrac}
\usepackage{microtype}
\usepackage{graphicx}
\usepackage{xcolor}
\usepackage{natbib}

\bibpunct{(}{)}{, }{n}{,}{,}
\hypersetup{colorlinks=true, linkcolor=blue, citecolor=blue, urlcolor=blue}
\newcommand{\figalttext}[2][]{}
\let\addedtable\relax

\title{Humans and AI respond similarly to task difficulty when reasoning about content moderation dilemmas}

\author{\href{https://orcid.org/0000-0002-5947-7490}{\includegraphics[scale=0.06]{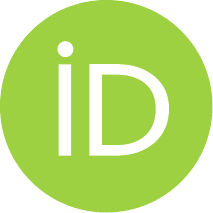}\hspace{1mm}Thomas R.~Davidson} \\
	Department of Sociology\\
	University of North Carolina at Chapel Hill\\
	\texttt{thomas.davidson@unc.edu}
}

\renewcommand{\headeright}{}
\renewcommand{\undertitle}{}
\renewcommand{\shorttitle}{Humans and AI respond similarly to task difficulty}

\hypersetup{
pdftitle={Humans and AI respond similarly to task difficulty when reasoning about content moderation dilemmas},
pdfauthor={Thomas R. Davidson},
pdfkeywords={large reasoning models, chain-of-thought, response times, content moderation, hate speech},
}

\begin{document}
\maketitle

\begin{abstract}
Large reasoning models (LRMs) generate intermediate reasoning steps before producing answers, improving performance on challenging problems by iteratively developing solutions. There is evidence that LRMs expend more effort on the problems that take people longer to solve, but it is unclear whether these relationships hold on more subjective tasks without verifiable answers. This study addresses this issue through an experiment using a content moderation task, where task difficulty is manipulated by varying the relative offensiveness of two different options. Comparing six large reasoning models with 1,854 human subjects, we find that models expend more reasoning effort, measured using the length of the chain-of-thought (CoT), on the kinds of cases that take people longer to adjudicate. People took 31\% longer and models produced 51 to 306\% longer CoTs when offensiveness was held constant across the two posts. The correlation is relatively low for individual decisions ($r = 0.15$--$0.19$), but average response times and average reasoning effort are closely correlated when tasks are grouped by the types of offensive language ($r = 0.85$--$0.93$). Qualitative analyses of the CoT summaries from Gemini 2.5 Pro corroborate these findings, showing that secondary factors were mentioned more often when offensiveness was fixed. The results point to the potential for using reasoning traces to prioritize difficult cases and enhance interpretability.
\end{abstract}


\paragraph{Significance statement}
Artificial intelligence systems are increasingly used to make judgments that have no single correct answer, in domains such as medicine, law, and hiring. Reasoning models work through a problem step by step before answering. If the effort they expend corresponds to what people find difficult, it could signal when decisions require further scrutiny. This study documents such a correspondence on a content moderation task: people took longer, and models reasoned for longer, when choosing between social media posts that were similarly offensive. In these cases, the reasoning we analyzed more often considered context, such as a post's topic and how others responded. Reasoning effort and content could help identify cases for human review and explain why they are difficult.

\section{Introduction}

Chain-of-thought (CoT) prompting---where large language models (LLMs) are instructed to think step-by-step before producing an answer---improves accuracy on various tasks \cite{wei_chain--thought_2022}. Large reasoning models (LRMs) are optimized to perform this intermediate reasoning before responding to user queries \cite{lightman_lets_2023,yao_tree_2023,guo_deepseek-r1_2025}. Computer scientists have drawn analogies to dual-process theories of cognition: whereas earlier LLMs responded immediately by generating the most likely token sequence (System 1), LRMs ``think'' before answering (System 2) \cite{kojima_large_2022,yao_tree_2023}. Beyond performance gains, the contents of the CoT represent the model ``thinking out loud'', aiding interpretability and AI safety efforts by increasing transparency \cite{guan_deliberative_2025,korbak_chain_2025}.

Recent work finds that the reasoning effort used by LRMs correlates with human response times (RTs) on tasks with objectively correct answers, such as arithmetic and logic problems, suggesting parallels in the allocation of effort \cite{de_varda_cost_2025}. Yet many of the consequential settings where LRMs are being deployed lack such ground truth. Tasks such as legal analysis, medical triage, and hiring all require judgments based on limited evidence, where reasonable people can reach different conclusions. In these domains, effort reflects not only computational complexity but also uncertainty about value-laden tradeoffs. There is also debate about whether any associations between human and LRM reasoning reflect meaningful similarities or are merely an artifact of training procedures \citep{dujmovic_no_2026,vankov_correlations_2026,hu_thinking_2026}. More generally, reasoning does not always improve performance \citep{illusion-of-thinking} and CoTs can be ``unfaithful'', obscuring information used to make decisions \cite{lanham_measuring_2023,turpin_language_2023,chen_reasoning_2025}. This study contributes to these debates by extending the evidence base to a more subjective task and assessing the extent to which CoT length and content can provide useful information about AI decision-making.

We analyze applications of LRMs to content moderation to examine whether the correspondence between AI reasoning effort and human cognitive effort extends to more subjective judgments. This application is not a hypothetical exercise since LRMs are already being deployed for content moderation. For example, OpenAI has released \texttt{gpt-oss-safeguard}, a model fine-tuned for moderation that uses reasoning to make decisions \cite{openai_technical_2025}. If the reasoning effort of these models tracks what people find difficult, it could provide a practical signal for content moderation, helping to identify more challenging cases to prioritize for human review. Moreover, if the CoT content provides insights into task difficulty, it may yield information that can help moderators to review challenging cases \cite{zhang_biasx_2023,zhou_cobra_2023}. Faithfulness is particularly important in this context given persistent biases in LLMs that may undermine the validity of automated decision-making if left undisclosed \cite{hofmann_ai_2024,bai_explicitly_2025}.

This study uses an experimental design to compare how LRMs and human subjects respond to content moderation dilemmas. The task consists of a forced-choice conjoint design \cite{hainmueller_causal_2014,hainmueller_validating_2015}, where pairs of synthetic social media posts that mimic online hate speech are shown and one must be prioritized for content moderation. This entails reviewing and evaluating the content of the messages and associated context such as the identity of the author and engagement metrics to reach a decision. An earlier study \citep{davidson-multimodal-2025} demonstrated similarities in how human subjects and vision-language models, which lacked reasoning capabilities, used specific features of the posts to make decisions. The present study implements new experiments with LRMs and analyzes the RTs from the same experiment, which were not used in the earlier study.

\begin{figure*}[t!]
\centering
\includegraphics[width=\textwidth]{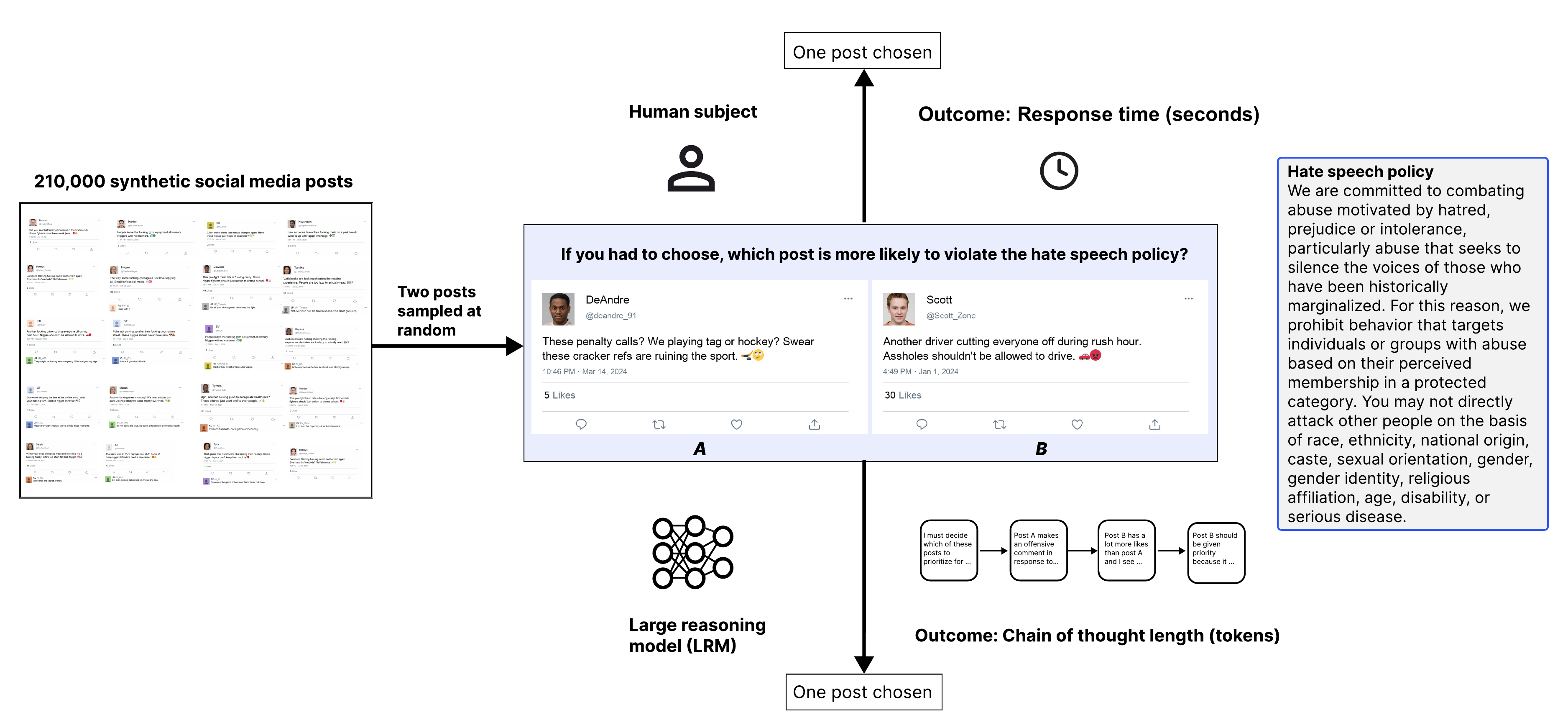}
\caption{Study design. A corpus of 210,000 synthetic social media posts was created, each varying according to the slur used, the presence of cursing, the topic, the race and gender of the user, whether there are positive or negative replies, and whether the number of likes is high or low. In the human subjects experiment, each subject was shown 15 pairs of posts sampled at random from the corpus and, for each pair, was asked to select which of the posts was more likely to violate the hate speech policy. The policy, on the right of the diagram, is based on that used by X (formerly Twitter). An example pair is shown in the blue box. The outcome analyzed in this study is the time in seconds it took the subject to make a decision, measured by the timing of the first click on the screen. The same pairs were input into each LRM along with a prompt, and the decision was recorded. The outcome is the length of the reasoning trace, measured using the number of tokens in the chain-of-thought.}
\label{fig:fig1}
\figalttext[Diagram of the study design.]{Diagram of the study design: synthetic posts varying in slur, cursing, user identity, replies, and likes are shown in pairs to human subjects and to reasoning models with a hate speech policy; human response time and the number of reasoning tokens are recorded.}%
\end{figure*}

We compare how human RT and LRM reasoning effort, measured using CoT length, vary according to the attributes in the posts evaluated. Specifically, we assess how both vary with the relative offensiveness of each pair of posts. We hypothesize that decisions will require less time for humans and shorter CoTs for LRMs when posts diverge in offensiveness. This expectation follows from work on bounded rationality, in which decision-makers adapt their effort to the structure of the task and develop satisficing strategies to reduce effort \citep{simon_rational_1956,payne_adaptive_1993}. When choosing between two options, a task can be simplified by deciding based on the most important attribute and consulting further attributes only when it does not discriminate between the options \citep{payne_adaptive_1993,gigerenzer_reasoning_1996}. In the context of the content moderation experiment, the presence of an offensive epithet in one post but not another provides a cue for distinguishing severity that can be used as a heuristic to quickly reach a judgment without weighing other information \cite{tversky_judgment_1974}, consistent with satisficing in surveys and experiments \cite{krosnick_response_1991,hainmueller_validating_2015}. In contrast, when offensiveness is either similar or held constant, decision-making requires more effortful deliberation over other information. In such cases, we expect longer RTs since the choice requires weighing more information \citep{hick_rate_1952,schneider_hicks_2018}. If LRMs exhibit a similar sensitivity to task difficulty, they will produce longer CoTs under such conditions. Unlike prior work that infers difficulty from observed RTs or accuracy, we manipulate difficulty ex ante by varying the language used in the posts, testing whether decisions require more effort when divergent offensiveness cannot be used as a heuristic.

The results of these experiments show consistent patterns across human subjects and each of the six models tested. Judgments involving posts with similar or identical perceived offensiveness took substantially longer and consumed more reasoning tokens than those with divergent levels of offensiveness. The correlations between RTs and CoT length at the decision level are small, consistent with substantial variation in RTs beyond that explained by the main effects of the manipulated attributes, but aggregation across similar stimuli reveals stronger correlations that are comparable in magnitude to those seen in objective tasks \citep{de_varda_cost_2025}. An analysis of the CoT summaries provided by Gemini 2.5 Pro shows that the reasoning traces mentioned secondary factors, such as replies and like counts, more often when both posts contained an identical slur. These findings are consistent with the proposed account of heuristic use and demonstrate how CoTs can provide quantitative and qualitative information about task difficulty. We discuss the implications of these findings for theoretical debates about reasoning models and the practical applications in content moderation and other tasks.

\section{Human response time and AI reasoning effort}

Reasoning models can perform better on various tasks than earlier LLMs by using additional compute at inference time \cite{openai_openai_2024,snell_scaling_2024}. These reasoning capabilities are developed through specialized post-training procedures that train models to perform CoT reasoning \citep{guo_deepseek-r1_2025,yang_qwen3_2025,team_kimi-vl_2025}. Some implementations rely on supervised fine-tuning on examples of tasks paired with reasoning traces \citep{muennighoff_s1_2025}, but reinforcement learning (RL) that rewards successful solutions to problems with objectively verifiable answers can elicit reasoning without such demonstrations \citep{guo_deepseek-r1_2025}. Open-weights LRMs, which have better-documented training procedures than closed-weights models, typically combine multiple phases of training, involving both supervised fine-tuning on examples and RL, as well as additional steps to improve efficiency \citep{guo_deepseek-r1_2025,yang_qwen3_2025,team_kimi-vl_2025}. CoT length---the number of intermediate reasoning tokens generated before an output is produced---provides a measure of the reasoning effort used to reach an answer or make a decision.

We investigate parallels between human RTs and LRM reasoning. RTs are widely used as proxies for cognitive phenomena ranging from attitude accessibility \citep{mulligan_response_2003} to implicit associations \citep{greenwald_measuring_1998}. We use RTs as a measure of the time taken to reach a decision, which we expect will increase in relation to task difficulty. Research in experimental psychology has consistently found that RTs increase with the complexity of the choice set and the amount of information that needs to be processed \citep{hick_rate_1952,bettman_componential_1990,payne_adaptive_1993,schneider_hicks_2018}. Of course, RT is not a measure of reasoning alone, as it bundles together perceptual encoding, reading, attention, deliberation, and motor execution \citep{bettman_componential_1990,mulligan_response_2003,hu_thinking_2026}. Moreover, there are idiosyncratic differences in how people respond to online surveys, with some ``speeding'' through questions and others carefully deliberating \citep{krosnick_response_1991,zhang_speeding_2014}. But while any individual RT may be noisy and RTs may vary systematically across subjects, aggregation across multiple tasks and subjects can help to isolate any underlying relationship between RTs and task difficulty.

Several studies analyze the relationship between human timing and the behavior of non-reasoning LLMs. Language model ``surprisal'' becomes a better predictor of human reading time as language model quality increases \citep{goodkind_predictive_2018}, but larger transformer-based models trained on more data diverge from this pattern \citep{oh_transformer-based_2023}. Research on logical reasoning finds a closer correspondence: language model \textit{confidence}---measured as the gap in log-probability between the top answer and the next-best option---is higher on items that humans also answer more rapidly \citep{lampinen_language_2024}. The same study also found that humans and LLMs both rely on semantic content when reasoning about novel logical problems, exhibiting parallel ``content effects'' that suggest heuristic use and shared sensitivities to task familiarity rather than purely abstract inference.

Reasoning models can enable a more direct comparison by measuring the amount of computational effort used to perform a task. de Varda et al. \cite{de_varda_cost_2025} document how CoT length closely tracks human RTs across seven reasoning tasks---numerical and verbal arithmetic, syllogisms, formal logic, relational and intuitive reasoning, and abstract structure mapping---correlating with average RTs at $r=0.34$ to $r=0.89$. Their primary analysis uses DeepSeek-R1 \citep{guo_deepseek-r1_2025}, and consistent patterns are present across other open-weights LRMs. Similar relationships are also observed on abductive reasoning tasks that require choosing between alternative hypotheses \citep{arthur_thinking_2026}. LRMs also use more reasoning tokens when processing association-incompatible information, consistent with research on implicit bias among human subjects \cite{lee_implicit-bias-like_2026}. These studies provide evidence that human RTs and LRM reasoning effort are correlated and that LRMs allocate more effort to difficult tasks. At the same time, there has been significant debate regarding the interpretation of these correlations \cite{de_varda_cost_2025}. We consider three main objections and outline how this study contributes to this emerging literature through a new evaluation on a more subjective task.%

First, correlations do not establish any connection at the algorithmic level between human and AI reasoning \citep{dujmovic_no_2026}, and related work points to the ways in which CoTs can deviate from human reasoning \citep{bao-etal-2025-likely}. However, it is important to emphasize that the aim of de Varda et al. \cite{de_varda_cost_2025} and the present study is not to show that humans and LRMs arrive at decisions via comparable internal processes, but rather to establish whether LRMs allocate effort in ways that track human RTs \citep{de_varda_reply_2026-2}. Systems that differ internally can still allocate effort in similar ways if each adapts to the task structure \citep{simon_rational_1956,anderson_adaptive_1990}. We respond to calls to explain why human RTs and LRM reasoning are correlated \citep{dujmovic_no_2026} by examining whether task difficulty increases effort for both humans and LRMs.

Second, associations between task difficulty and CoT length could reflect patterns learned during training, where more difficult problems are paired with longer demonstrations, rather than adaptive effort allocation \citep{vankov_correlations_2026} (for a response, see \citealp{de_varda_reply_2026-1}). However, this account only partially reflects how LRMs are trained. Many contemporary LRMs, including DeepSeek-R1, are also trained with RL that rewards successful solutions, so reasoning length can be shaped by reward-based optimization as well as by imitation \citep{guo_deepseek-r1_2025,yang_qwen3_2025,team_kimi-vl_2025}. Since these rewards are mainly defined for problems with verifiable answers, it is an open question whether the allocation of effort generalizes to tasks without them. We consider a novel task where there is no correct answer, all inputs are similar in length,\footnote{For some models there is minor variation in the number of input tokens used for different problems. We control for input token count in these cases to address this issue.} and difficulty is manipulated ex ante. Under these conditions, it is less plausible that longer traces can be attributed to longer inputs or to memorized responses to these particular stimuli \citep{vankov_correlations_2026}. LRMs may nonetheless have learned a general tendency to reason at greater length about ambiguous comparisons, which would not make CoT length less informative about which decisions are difficult.

\begin{figure*}[t!]
\centering
\includegraphics[width=\textwidth]{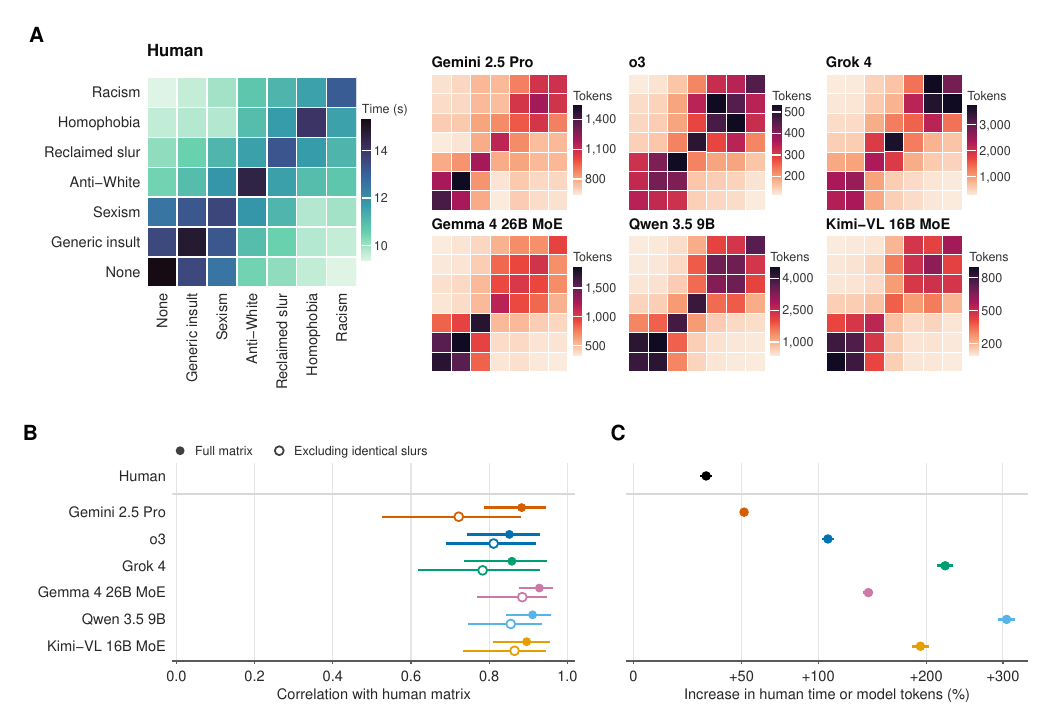}
\caption{Human response time and AI reasoning effort across combinations of slurs. (A) The large heatmap on the left shows the mean RT for each pair of slurs, in seconds, averaged on the log scale after adjustment for subject fixed effects and the log task number. The six smaller heatmaps on the right show the mean number of reasoning tokens, averaged on the log scale, consumed by each closed-weights model (top row) and open-weights model (bottom row), with the same axes as the human heatmap (see Fig.~S3 for labeled pair-level estimates). (B) Correlation between the human matrix and each model's matrix, using the mean adjusted log RT and the mean log reasoning tokens for all pairwise combinations (filled) and for the 21 combinations on the off-diagonal, excluding the identical slurs (open). Error bars represent 95\% bootstrap confidence intervals. (C) Percentage increase in human RT and in each model's reasoning tokens when both posts contain the same slur, on a log scale. The human model includes subject fixed effects and the log task number, with standard errors clustered by subject. Error bars represent 95\% confidence intervals. All panels use the analytic sample of $N = 27{,}071$ pairs ($N = 27{,}052$ for o3).}\label{fig:fig2}
\figalttext[Heatmaps and dot plots of human response time and model reasoning tokens across pairs of slurs.]{Seven heatmaps across pairs of slur categories, a large one of mean adjusted human response time and six small ones of mean reasoning tokens for the six models, all darkest along or near the diagonal; a dot plot of correlations between each model's matrix and the human matrix, between 0.72 and 0.93; and a dot plot of the percentage increase in time or tokens for same-slur pairs.}%
\end{figure*}

Third, CoTs might be ``performative'', containing tokens that are decoupled from the internal computational effort \citep{hu_thinking_2026}. More generally, the content of reasoning traces may not provide a faithful record of the decision-making process \citep{lanham_measuring_2023,turpin_language_2023,chen_reasoning_2025}. However, reasoning traces need not be faithful for their length to track task difficulty \citep{de_varda_reply_2026}, and we assess whether CoT length responds to a manipulation of difficulty that is defined ex ante. While we cannot test faithfulness directly, by analyzing the content of the CoT summaries, we evaluate whether the factors they discuss vary with task difficulty in ways that are consistent with our account.

Language models---with and without additional reasoning---can make judgments that are consistent with those made by people \cite{loru_simulation_2025,bhatia_computational_2025,binz_foundation_2025} and can exhibit human-like cognitive biases \cite{lampinen_language_2024,cheung_large_2025,lee_implicit-bias-like_2026}. However, we know little about whether the effort expended in reaching decisions on more subjective tasks is correlated with human RTs given that LRMs are primarily optimized on tasks with verifiable answers \citep{guo_deepseek-r1_2025}. We conduct, to our knowledge, the first direct comparison between LRM and human decision-making on a task that simulates subjective choices in a practical setting using multimodal stimuli. \autoref{fig:fig1} summarizes the research design, which builds on an earlier human subjects experiment \citep{davidson-multimodal-2025}. A diverse pool of human subjects ($N = 1{,}854$) each chose which post to prioritize for violating a hate speech policy across 15 pairs sampled from a corpus of 210,000 synthetic social media posts ($N = 27{,}810$ decisions), and their RTs were recorded. The same task was then given to three closed-weights LRMs (OpenAI's o3, Google's Gemini 2.5 Pro, and xAI's Grok 4) and three open-weights models (Google's Gemma 4 26B MoE, Moonshot AI's Kimi-VL 16B MoE, and Alibaba's Qwen 3.5 9B), and the CoT length was recorded for each decision. These models differ in training data, size, and architecture, enabling us to examine the extent to which any patterns generalize across LRMs.

\section{Reasoning about content moderation dilemmas}

We first examine the patterns of human RT and AI reasoning effort as a function of the language evaluated in each content moderation task. The heatmaps in \autoref{fig:fig2}A depict the distribution of effort across the combinations of slurs in each pair of posts. The left heatmap shows the mean adjusted RT for each combination, averaged on the log scale, where terms are ordered by perceived offensiveness among human subjects, ranging from no slurs and generic insults without identity-based connotations at the lower end to homophobia and racism at the higher end \cite{davidson-multimodal-2025}. The right panel shows the mean CoT length for each of the six models, averaged in the same way. The same general pattern is present across all heatmaps: the color is darkest along the diagonal and in adjacent cells, indicating that decisions required the most time from human subjects and the most reasoning effort from the LRMs when the language across both posts was either identical or similar in perceived offensiveness. 

Human subjects took the longest to make decisions when neither post contained offensive terms or both had relatively low perceived offensiveness. They responded fastest when shown divergent pairs, such as homophobic or racist terms paired with more innocuous language, consistent with our expectations. Similar patterns are evident across all six LRMs. The decisions required the least reasoning when the offensiveness was most divergent, specifically when racist or homophobic epithets were paired with posts without slurs or with generic insults (see the top-left/bottom-right corners of each heatmap). Each model exhibits clustering along the diagonal but has a different ``signature'' across different terms, pointing to idiosyncrasies in how each model reasons about various kinds of speech. Among the closed-weights models, Gemini had the longest CoTs when comparing posts that both featured a generic insult, o3 expended more effort when adjudicating between homophobia and reclaimed anti-Black slurs as well as sexism and generic insults, and Grok's reasoning was greatest when comparing pairs of anti-White posts or deciding between racism and homophobia. The open-weights models show similar patterns (\autoref{fig:fig2}A, bottom row), with higher effort when evaluating posts with comparable or identical levels of offensiveness.

To assess the overall similarity in how humans and LRMs responded to variation in task difficulty, we measured the correlations between the matrices. \autoref{fig:fig2}B shows that overall correlations are strong across all six models, ranging from $r = 0.85$ to $0.93$. To assess whether this holds without the identical pairs, we repeated the analysis excluding the diagonal entries, finding correlations ranging from $r = 0.72$ to $0.88$ (all $p < 0.005$ using permutation tests; Table~S4). This shows that the similarity does not arise only from the pairs with identical slurs, but also holds across pairs whose slurs differ.

To quantify the degree to which task difficulty impacts effort, we analyzed differences in RTs and CoT length as a function of the relative offensiveness of the two posts. At a high level, we assess the contrast between cases when both posts contain the same slur and other examples, comparing the diagonal versus off-diagonal cells of the heatmaps. \autoref{fig:fig2}C shows that human subjects took approximately 31\% longer ($\hat{\beta}=0.273, p<0.001$) to make decisions in these cases, and all six LRMs expended substantially more reasoning effort, ranging from a 51\% increase in tokens for Gemini 2.5 Pro ($\hat{\beta{}}=0.415, p<0.001$) to a 306\% increase for Qwen 3.5 9B ($\hat{\beta{}}=1.400, p<0.001$). We also tested whether these relationships hold at a more granular level by estimating regression models with dummy variables for each pair, using posts with the greatest offensiveness disparity (None $\times$ Racism) as the reference category. These results (see Fig.~S3 and Table~S11) show substantial differences in effort in response to the relative offensiveness of the two posts shown. Large effects were observed even when both posts lacked slurs entirely (None $\times$ None). Humans took approximately 69\% longer ($\hat{\beta{}}=0.525, p<0.001$) and LRMs ranged from using 109\% more reasoning tokens for Gemini 2.5 Pro ($\hat{\beta{}}=0.736$) to 1{,}009\% for Qwen 3.5 9B ($\hat{\beta{}}=2.406$); all $p<0.001$. Pairing slurs of comparable but distinct severity also yielded substantial increases in reasoning effort. For example, Grok 4 used over ten times as many tokens on Homophobia $\times$ Racism pairs ($\hat{\beta{}}=2.474, p<0.001$) compared to when a racist post was shown alongside one without a slur, while human subjects took 26\% longer on these cases ($\hat{\beta{}}=0.230, p<0.001$). 

\begin{figure}[t!]
\centering
\includegraphics[width=0.7\linewidth]{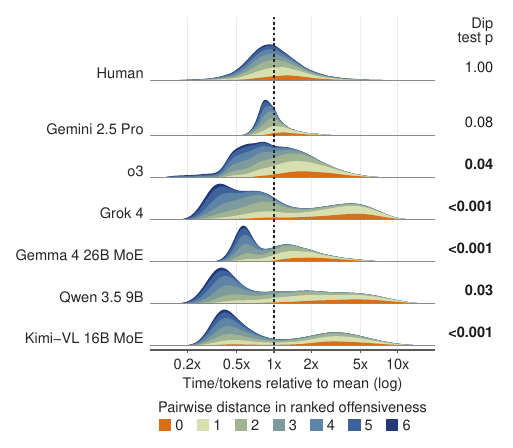}
\caption{Density of human RT and AI reasoning effort, by the distance between the slurs. Each row shows the distribution for humans (top row) or LRMs (rows 2--7), relative to the row mean on the log scale. The vertical dashed line marks the mean (1x). The layers represent the distance between the two slurs in the offensiveness ranking, from 0 at the bottom, when both posts contain the same term, to 6 at the top, for None $\times$ Racism. Human RT is adjusted for subject fixed effects and the log task number. o3's reasoning tokens, reported in blocks of 64, are spread uniformly within each block for comparability. The right column gives the $p$-values from Hartigan's dip test, with values below 0.05 in bold, indicating a rejection of the null hypothesis of unimodality. $N = 27{,}071$ pairs ($N = 27{,}052$ for o3).}
\label{fig:dist}
\figalttext[Stacked density curves of human response time and model reasoning tokens.]{Seven stacked density curves, one for humans and one for each of six models, on a shared axis from 0.1 to 20 times each row's mean, with a dashed vertical line at the mean. Layers are colored by the distance between the two slurs, from orange for the same slur at the bottom, through pale yellow-green for adjacent slurs, to navy for slurs far apart at the top. The human curve has a single peak in which the layers overlap. Most model curves have two peaks, with far-apart pairs in the low-effort peak, left of the line, and same or adjacent pairs in the high-effort peak, right of it. A column gives the dip-test p-values: 1.00 for humans, 0.08 for Gemini 2.5 Pro, and below 0.05 for the other five models.}%
\end{figure}

\autoref{fig:dist} shows the distributions of RTs and reasoning tokens and how they vary according to the relative offensiveness of the posts in each task, illustrating the overall relationships between effort and difficulty. In the top row, human RTs tend to be longer when the distance in offensiveness is lower. There is greater variability in responses across the models, with decisions for distant pairs requiring considerably less reasoning effort than average, and those with similar or adjacent levels of offensiveness sometimes requiring five to ten times as much effort, consistent with the previous regression analyses. Whereas the human distribution appears approximately log-normal, several models seem to have bimodal distributions. To test whether the distributions diverge from unimodality, we conducted Hartigan's dip test \citep{hartigan_dip_1985}, calculating $p$-values from 2{,}000 simulated replications. The test did not reject unimodality for adjusted human RTs ($p = 1.00$), but the distributions of five of the six models departed from unimodality ($p < 0.05$; Gemini 2.5 Pro, $p = 0.08$). This is consistent with models' effort varying more sharply with difficulty than human RT.

\section{Assessing the correlations between human response time and AI reasoning effort}

\begin{figure*}[t!]
\centering
\includegraphics[width=\textwidth]{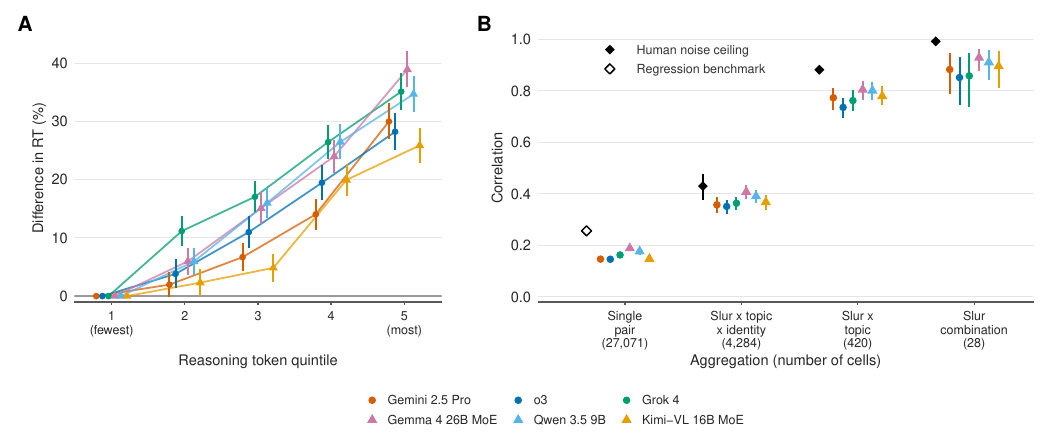}
\caption{Associations between reasoning effort and human response times. (A) Percentage difference in adjusted human RT for each quintile of reasoning tokens, relative to the bottom quintile, estimated using OLS with subject fixed effects and controls for the log task number (Table~S9). Error bars represent 95\% confidence intervals. (B) Pearson correlation between adjusted log RT and log reasoning tokens across design attributes, from single pairs to the aggregate slur combinations. The single-pair estimates are correlations across individual pairs, with RT adjusted for subject fixed effects and the log task number. The open diamond (``Regression benchmark'') denotes the correlation with single decisions that a predictor capturing the main effects of all manipulated attributes would reach (see Materials and methods). The other three sets of estimates show, for each model, the correlation between the mean adjusted log RT and the mean log reasoning tokens across cells of pairs that share a combination of attributes, with the cell count shown in parentheses. Filled diamonds mark the human noise ceiling, the highest correlation any predictor could reach given the noise in the human cell means, estimated from random split halves of each cell with the Spearman--Brown correction (Table~S3). The slur-combination level is the same correlation across all 28 combinations shown in \autoref{fig:fig2}B. Error bars represent 95\% bootstrap confidence intervals (see Materials and methods). $N = 27{,}071$ pairs ($27{,}052$ for o3).}
\label{fig:fig3}
\figalttext[Line chart and dot plot of the association between reasoning effort and human response time.]{A line chart showing human response time rising steadily across the quintiles of pairs ranked by each of six models' reasoning tokens, to 26 to 39 percent longer in the top quintile than in the bottom quintile; and a dot plot of the correlation between reasoning effort and human response time at four levels of aggregation, which rises from about 0.15 for single pairs to about 0.4, 0.75, and 0.9 as pairs are averaged within design cells, just below a human noise ceiling of about 0.43, 0.88, and 0.99, with a regression benchmark of about 0.26 for single pairs.}%
\end{figure*}

We now more closely consider the relationship between human RT and LRM effort. We estimated regression models predicting log RT as a function of log CoT length for each LRM, accounting for subject-level variation and task-ordering effects that otherwise partially suppress the relationship (see Fig.~S2). The estimated elasticities are positive and statistically significant for all six models: a 1\% increase in CoT length is associated with between 0.09\% and 0.28\% longer RTs, depending on the model (all $p < 0.001$; Table~S7). This establishes that, on average, the tasks that took longer for human subjects to decide required more reasoning effort. To better quantify the relationship, we estimated a second model predicting log RTs relative to quintiles of CoT length for each LRM. \autoref{fig:fig3}A shows that subjects took 26--39\% longer on the fifth of pairs where a model reasoned the most than on the fifth where it reasoned the least. For all six models, the estimated RT rises from each quintile to the next, showing that the relationship is not driven only by the extremes (see Table~S9 for full estimates).

\autoref{fig:fig3}B shows how the correlations between human RTs and CoT length vary across different aggregations. The overall correlation, shown on the left, is relatively low, ranging from $r = 0.15$ to $0.19$ ($0.11$--$0.14$ unadjusted; Fig.~S2). To put this into context, we used a regression benchmark model to estimate adjusted RT as a function of the attributes manipulated in each task. We find that the main effects of the manipulated attributes explain about 6.6\% of the variance in adjusted log RT (SI, ``Correlations of averaged decisions''), which implies that a predictor capturing these effects correlates with single RTs at about $r = 0.26$. Much of the variation thus remains unexplained by these main effects, consistent with RTs bundling reading, attention, deliberation, and motor execution. This helps to contextualize the weak individual-level correlations, although the benchmark does not capture interactions among the attributes. By averaging over pairs that share design attributes, however, we can remove much of the idiosyncratic variation and more closely assess parallels in effort due to task difficulty. When pairs are grouped by the combination of slurs, topics, and user identities in the two posts (4,284 cells of approximately six pairs each), the correlations range from $r = 0.35$ to $0.41$. Grouping by slurs and topics raises them to $0.74$--$0.80$ (420 cells), and grouping by slurs alone shows high correlations ($0.85$--$0.93$), identical to the full-matrix similarity analysis in \autoref{fig:fig2}B. At each of these levels, the correlations approach the human noise ceiling---the highest correlation any predictor could reach given the noise in the human averages---estimated following the split-half method used in prior work \citep{de_varda_cost_2025,arthur_thinking_2026}. The ceiling rises to $0.43$, $0.88$, and $0.99$, respectively, as averaging removes noise (filled diamonds; Table~S3). In aggregate, CoT length closely tracks human RTs across kinds of decisions. The increase in correlations does not arise from averaging alone. For each grouping, we conducted a placebo analysis, reassigning pairs at random to groups of equivalent size that did not depend on the manipulated variables and recalculating the averages. In all three cases, the resulting correlations are similar in magnitude to the single-pair values (mean $r = 0.14$--$0.19$; Table~S3, Panel B).

\section{Analyzing AI reasoning using chain-of-thought summaries}

\begin{figure*}[t!]
\centering
\includegraphics[width=0.9\textwidth]{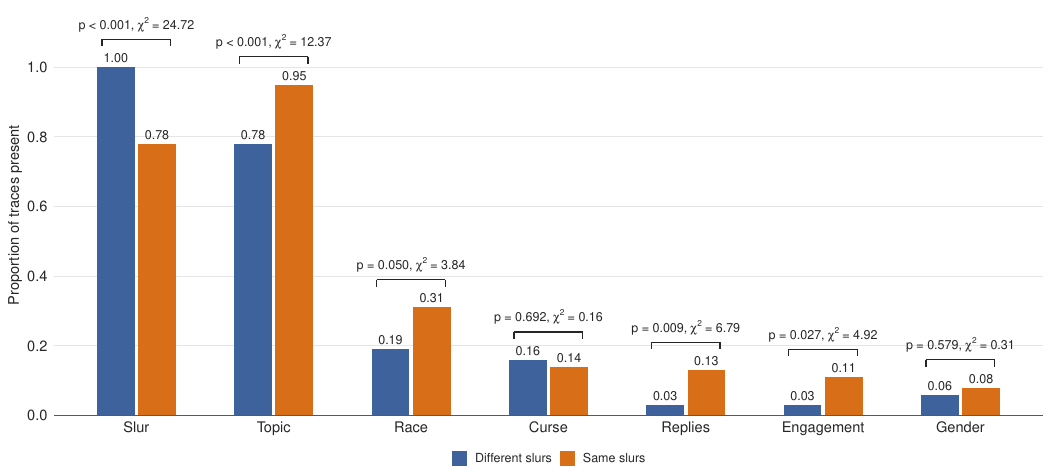}
\caption{Proportion of chain-of-thought summaries that mention each factor manipulated in the experiment. The bars show the proportion of reasoning traces from Gemini 2.5 Pro that were coded as mentioning each factor manipulated in the posts, comparing pairs of posts where the slurs were different and pairs of posts where the slurs were constant ($n=200$). Two-sided chi-squared tests (df $=$ 1) compare the proportions across the two groups, with $p$-values and test statistics reported above each category.}\label{fig:fig4}
\figalttext[Bar chart of the factors mentioned in CoT summaries.]{Bar chart of the share of Gemini 2.5 Pro CoT summaries mentioning each manipulated factor, comparing different-slur and same-slur pairs. Slurs are mentioned more when CoTs describe cases with different slurs. Topic, race, replies, and engagement are more frequently mentioned in CoTs for pairs with the same slurs.}%
\end{figure*}

We conducted a content analysis of the CoT summaries from Gemini 2.5 Pro---the only closed-weights model to provide this information at the time of the analyses---to understand how the model described its reasoning about these decisions. We focused on whether each factor manipulated in the experiment was mentioned in each CoT. Three raters coded a random sample of 200 summaries, where half were pairs where both posts used the same slur and the other half used different terms (see Materials and methods). The frequencies across the two groups are shown in \autoref{fig:fig4}, and complete examples of the summaries are provided in the SI (``Reasoning traces'').

When offensiveness diverged, Gemini tended to prioritize language generally perceived as more offensive \cite{davidson-multimodal-2025}, and typically gave a brief explanation. These explanations mentioned the slur in every case coded, compared with 78\% of cases when offensiveness was constant ($\chi^2(1) = 24.72$, $p < 0.001$), but this difference is largely explained by the fact that the latter condition includes twenty pairs where neither post contained any offensive term (None $\times$ None). When slurs were held constant, the reasoning traces were significantly more likely to mention several contextual factors. The post topic was frequently mentioned across both groups, but more often in these cases (95\% compared with 78\%; $\chi^2(1) = 12.37$, $p < 0.001$). In some cases, the topic was used as the primary justification for the decision. For example, Gemini mentioned selecting political posts over those about sports or everyday issues. The race of the users---which the model inferred based on the names and profile images---was also mentioned more often, although the statistical significance is borderline ($\chi^2(1) = 3.84$, $p = 0.050$). When an anti-Black reclaimed slur was used, Gemini sometimes discussed how the term took on a different meaning in African American Vernacular English, and prioritized posts by White users. The presence of replies to the original post ($\chi^2(1) = 6.79$, $p < 0.01$) and the number of likes ($\chi^2(1) = 4.92$, $p < 0.05$) were also mentioned more frequently in these cases. For example, when comparing two posts using racial slurs, Gemini provided the following explanation: ``Engagement is critical here. Image A has 2 likes. Image B has 26 likes \textit{and} a reply in agreement. That level of engagement is a red flag. Higher engagement equals higher potential harm. More people are seeing it, and it's being amplified by the reply.'' These analyses reveal how the model navigated ambiguity when the primary cue was unavailable by considering contextual factors, such as the topic of discussion, race, and audience reach as a proxy for potential harm.\footnote{We did not systematically analyze the CoT summaries from the open-weights models, but their content appeared qualitatively similar, often describing the task and mentioning various aspects of the posts.}

\section{Discussion}

This study documents a correspondence in how humans and large reasoning models allocate effort in response to difficulty across content moderation decisions. People took longer and LRMs expended more effort when the two posts shown had similar or identical levels of offensiveness. The correlations for single decisions are relatively low, consistent with substantial variation in adjusted log RT beyond that explained by the main effects of the manipulated attributes. The correlations across kinds of decisions are strong and approach the human noise ceiling. These results extend prior findings from formal reasoning domains \cite{lampinen_language_2024,de_varda_cost_2025,arthur_thinking_2026}, showing similar patterns on a task with no correct answer---the kind of subjective, value-laden judgment where LRMs are increasingly deployed in practice. The aggregate correlations are comparable in magnitude to those reported for problems with verifiable answers, where reasoning tokens correlated with the average RT of the subjects who solved each problem at $r = 0.34$--$0.89$ within tasks \cite{de_varda_cost_2025}.

These findings contribute to the debates about the relationship between human and AI reasoning, addressing calls for testable hypotheses about why RTs and reasoning effort are associated \citep{dujmovic_no_2026}. Content moderation involves interpretation and judgment, and we hypothesized that both humans and LRMs would expend less effort when divergent offensiveness could be used as a heuristic. The results are consistent with these expectations, showing relationships between task difficulty and effort across six models that vary in architecture, training data, and post-training procedures. These findings suggest that models trained to reason through problems with verifiable answers can generalize this behavior to a task without them. Since the posts share a common format, differ only in the manipulated attributes, and are distinct from the verifiable problems typically used to train LRMs to reason, it is less plausible that the CoT length is simply a function of longer inputs or memorization \citep{vankov_correlations_2026}. The qualitative analyses provide further support for our account of task difficulty and demonstrate the value of directly analyzing the content of the reasoning traces. When offensiveness was held constant, Gemini's reasoning traces mentioned secondary, contextual factors more often, consistent with theoretical expectations regarding satisficing and heuristics \citep{simon_rational_1956,tversky_judgment_1974,payne_adaptive_1993,gigerenzer_reasoning_1996,lampinen_language_2024}. While we cannot eliminate the possibility that these are post hoc rationalizations or establish whether they are faithful to the underlying computational process \citep{hu_thinking_2026}, we found that the CoT summaries contained plausible rationales that were often consistent with our own assessments of difficulty and mentioned variables determined to be important in related work \citep{davidson-multimodal-2025}.

Alongside these similarities, our study draws attention to differences between human and LRM reasoning effort. The effort distribution had a greater range among LRMs than humans: same-slur pairs took people 31\% longer than other pairs, but increased the models' reasoning tokens by 51--306\%. These disparities may reflect a tendency for LRMs to ``overthink'' the problems \cite{illusion-of-thinking}, whereas humans tend to limit effort once a choice is good enough \citep{simon_rational_1956}. However, the differences in these measures alone do not establish whether the additional reasoning was unnecessary. In terms of the overall distributions, adjusted human RTs appeared unimodal, whereas CoT lengths of most models (except Gemini 2.5 Pro) appeared to be bimodal. Reasoning was brief when the slurs were far apart in offensiveness and extended when they were the same or adjacent (\autoref{fig:dist}). To some extent, these patterns resemble the distinction between fast and slow thinking that has been used to describe LRMs \citep{kojima_large_2022,yao_tree_2023,lampinen_language_2024}. However, all models used additional reasoning, and a single process applied to a mix of easier and harder pairs could also produce a bimodal distribution. As such, it is unclear whether these disparities in reasoning effort reflect distinct processes or different positions on a continuum, paralleling debates about human cognition \citep{sloman_empirical_1996,stanovich_individual_2000}. Further work is needed to understand differences across models and the extent to which theories of human cognition provide a useful framework for explaining these processes.

These findings have practical implications for content moderation and other high-stakes settings where LRMs are deployed. Since CoT length tracks the kinds of decisions that people find difficult, reasoning effort could serve as a signal for cases that warrant human review or closer scrutiny. Such a signal is likely to be most reliable for identifying kinds of cases, where the correspondence with human RTs is strong. In supplementary analyses (Fig.~S4), we found that models usually made the same choice as the subject on easier tasks, selecting the same post in 84--97\% of pairs when slurs diverged in rank by the greatest margin. In such cases, both LRMs and humans chose the post with the more offensive term the vast majority of the time. Agreement decreased as the offensiveness gap narrowed and was substantially lower on the pairs where models reasoned most: LRMs chose the same post as human subjects on 51--63\% of pairs in the top quintile of CoT length, compared with 75--95\% of those where they reasoned least. It is important to emphasize that lower agreement does not indicate that additional reasoning leads to worse choices, since these tasks have no correct answer and each pair was judged by only one subject, so we cannot measure any consensus positions. Together with the longer human RTs on pairs with similar offensiveness, these results suggest that CoT length may also help to surface individual decisions where a model's choice is less likely to match a person's and more careful deliberation is required to adjudicate. Building on work on automated explanations \cite{zhang_biasx_2023,zhou_cobra_2023}, reasoning traces may also offer useful explanations to assist human moderators by providing information regarding why some decisions are difficult. For example, the discussion of reclaimed slurs in Gemini's reasoning traces shows sensitivity to the user identity, which might help to avoid a common type of false positive in hate speech detection \cite{sap_risk_2019,davidson_racial_2019}. Beyond content moderation, reasoning effort and traces may enhance the transparency of decision-making in domains such as law, medicine, and hiring, where AI systems are increasingly deployed, by providing signals for task difficulty and interpretable rationales for automated decisions \cite{korbak_chain_2025,brodeur_performance_2026}.

The study has several limitations. It focuses on a single content moderation task, so additional research is needed to assess whether these patterns generalize to other domains, including other tasks involving subjective judgments that have been used to compare human and AI responses \citep{bhatia_computational_2025,cheung_large_2025}. The experimental design entails some trade-offs between internal and external validity. The forced-choice format is widely used in conjoint experiments as it elicits comparative reasoning over complex desiderata \citep{hainmueller_causal_2014,hainmueller_validating_2015}. While content moderation workflows typically involve a series of individual judgments, comparisons are still useful insofar as these judgments are not entirely independent and triage entails prioritization and ranking \citep[for further discussion, see][]{davidson-multimodal-2025}. Further work is needed to establish whether these patterns hold on tasks where content is evaluated in isolation. The format of the posts is intended to have ecological validity while maintaining experimental control, but existing content is more varied and complex than the posts considered here. Supplemental analyses comparing the texts to an existing corpus of tweets annotated for hate speech show that they are somewhat longer, use longer words, and have a slightly higher reading grade (Table~S2). Finally, the human and AI responses follow different sampling processes: the human results pool decisions across multiple subjects, whereas each model rates every pair of posts.

Building on these findings, further work is needed to explain the processes underlying the correspondence reported here \citep{dujmovic_no_2026,de_varda_cost_2025,de_varda_reply_2026-2}. Research on human decision-making decomposes effort into elementary operations, such as reading and comparing pieces of information \citep{bettman_componential_1990}. An LRM's CoT is likewise the output of many internal computations over successive tokens. Measuring these operations directly, through process tracing for humans and analysis of complete reasoning traces for LRMs, could provide more clarity on which aspects of these processes are most closely associated. Controlled comparisons of models trained from the same base model with and without reinforcement learning would provide a more direct way to assess whether the correspondence depends on imitation of demonstrations or also arises under reward-based training \citep{vankov_correlations_2026,guo_deepseek-r1_2025,yue_does_2026}. Closer insight into the models' internal computations is also needed to determine whether the explanations in the reasoning traces reflect how the models actually arrived at decisions \cite{turpin_language_2023,chen_reasoning_2025,loru_simulation_2025}. While reasoning can improve performance on verifiable tasks, additional research is needed to understand whether reasoning effort improves decision quality, which remains an open question in domains without clear ground truth. We know that LLMs can make decisions aligned with those from humans \citep{bhatia_computational_2025,davidson-multimodal-2025}, but it is unclear how much these benefit from reasoning, or whether reasoning can reduce the prevalence of cognitive biases and misaligned behaviors \citep{cheung_large_2025,binz_foundation_2025}. For content moderation, comparisons against a plurality vote or an expert consensus could assess whether additional effort is associated with better decisions.

\section{Materials and methods}

\subsection{Conjoint task}

The experiment consists of a forced-choice conjoint task involving comparisons across pairs of synthetic social media posts \citep{hainmueller_causal_2014}. The offensiveness of each post was manipulated by modifying a template by adding different epithets, or slurs, such that posts varied in offensiveness from innocuous material without any slurs to those using racist, sexist, and homophobic language. Each post also varied across additional attributes that were randomized independently: user race and gender, cursing, topic, and engagement metrics. A total of 210,000 unique posts were created, representing all possible combinations of attributes, each of which was stored as an image file resembling a screenshot. 

\subsection{Human subjects experiment}
Adults in the United States ($N=1{,}854$) were recruited via Prolific, an online opt-in panel provider. Subjects who consented to participate were shown 15 pairs of posts on Qualtrics, sampled at random from the corpus, and a content policy resembling that used by X (formerly Twitter). In each case, subjects were asked to select which post was more likely to violate the policy. The decisions were recorded, along with the time it took for subjects to click on the screen for each task. The experiment is described in further detail in a related paper comparing the human responses to those from multimodal models (without reasoning capabilities) \citep{davidson-multimodal-2025}. This study analyzes the timing information, which was not used previously. See the SI for additional information (``Conjoint design and human subjects experiment'').

\subsection{LRM experiments}

\begin{table*}[t!]
\centering
\caption{\textbf{Models used in the LRM experiments.} This table provides information on each of the models used in the experiment. Release dates correspond to the public launch of each model. Parameter counts for the closed-weights models are not publicly disclosed. For the two mixture-of-experts (MoE) open-weights models, the total parameter count is shown along with the active parameter count in parentheses. Reasoning tokens are the mean and standard deviation (in parentheses) per pair across all 27{,}810 decisions (27{,}791 for o3).}
\label{tab:model_comparison}
\setlength{\tabcolsep}{3.5pt}
\begin{tabular}{l l c c l r}
\toprule
\textbf{Model} & \textbf{Developer} & \textbf{Released} & \textbf{Open-weights} & \textbf{Parameters} & \textbf{Reasoning tokens} \\
\midrule
Gemini 2.5 Pro    & Google DeepMind & 2025-03 & $\times$    & Undisclosed & 897.9 (419.6) \\
o3                & OpenAI          & 2025-04 & $\times$    & Undisclosed & 303.5 (241.5) \\
Grok 4            & xAI             & 2025-07 & $\times$    & Undisclosed & 1{,}600.3 (1{,}821.9) \\
Kimi-VL 16B MoE   & Moonshot AI     & 2025-06 & \checkmark & 16B (3B active, MoE) & 414.3 (516.5) \\
Qwen 3.5 9B       & Alibaba         & 2026-03 & \checkmark & 9B & 2{,}101.4 (2{,}572.7) \\
Gemma 4 26B MoE   & Google DeepMind & 2026-04 & \checkmark & 26B (3.8B active, MoE) & 780.5 (624.1) \\
\bottomrule
\end{tabular}
\end{table*}

We replicated the experiment across six LRMs, described in \autoref{tab:model_comparison}. Each model was presented with the same pairs of images shown to human subjects and instructed to choose which image was more likely to violate the policy. The response and the token usage were recorded. The closed-weights models were three of the top-performing frontier LRMs at the time of the experiment that could handle visual and textual data. For each model, the relevant application programming interface (API) was used to pass a prompt containing instructions and each pair of images (see SI, ``Prompt and LRM experiments''). The open-weights models were three of the most capable available at the time of the analyses and were served via vLLM on a high-performance computing cluster, using four NVIDIA L40S GPUs for each model. Further details on the setup are provided in the SI (``Open-weights models'').

\subsection{Variables}
Task difficulty is defined ex ante by the experimental manipulation---the slurs randomly assigned to the two posts---rather than inferred from observed RTs or CoT length. The seven slur conditions are ordered by offensiveness ratings from prior work, from posts without slurs to racist language \citep{davidson-multimodal-2025}. The distance measure in \autoref{fig:dist} is the rank difference between the two terms shown in each pair.

The outcome variable from the human subjects experiment is the time in seconds from when a pair of posts is presented until the first click on the screen, recorded by Qualtrics. This first-click time is the best available proxy for the RT between a subject being shown a pair and making a choice. The full dataset consists of 27,810 decisions by 1{,}854 subjects. A small number of outliers were removed prior to the analyses, namely instances where subjects took excessively long (and were likely inattentive) or completed the task unreasonably quickly. Specifically, cases where subjects took over 120 seconds or answered in one second or less were dropped, along with three cases from subjects left with a single pair ($N = 739$ pairs in total). The final analytic sample consists of $N = 27{,}071$ pairs of posts. Excluding outliers, the mean and median times are 15.3 and 11.4 seconds per pair, respectively. 

Following prior work \citep{lampinen_language_2024,de_varda_cost_2025,arthur_thinking_2026}, we use the natural logarithm of RTs and of reasoning tokens to account for their right-skewed distributions and for differences in scale across LRMs. To account for differences in RTs across subjects and task order,
we estimate the following regression:

\begin{equation}
\log(RT_{ij}) = \alpha_i + \beta \log(\tau_{ij}) + \varepsilon_{ij},
\end{equation}

\noindent where $RT_{ij}$ is the response time for subject $i$ on task $j$,
$\alpha_i$ represents subject fixed effects, and $\tau_{ij}$
denotes the task order within subjects (ranging from 1 to 15).
We then construct the adjusted log RT by adding the grand mean of log RT back to the estimated residuals:

\begin{equation}
\log(RT_{ij})^{\mathrm{adj}} = \hat{\varepsilon}_{ij} + \overline{\log(RT)}.
\end{equation}

\noindent This adjustment removes between-subject differences and systematic effects of task order while preserving the overall mean log RT. The adjusted log RT is used in the heatmaps, distributions, and correlation analyses.

The primary variable from the AI experiments is the number of reasoning tokens consumed by each model for a given task. Token counts are not directly comparable across models, which differ in architecture and use different tokenizers to split text and images into units; the mean and standard deviation for each model are shown in \autoref{tab:model_comparison}. For o3, the number of reasoning tokens was always reported in blocks of 64; 19 pairs without reasoning tokens were excluded from the analysis.

\subsection{Regression models}

The difference between same- and different-slur pairs was estimated using ordinary least squares (OLS) regression. For humans, log RT was regressed on a dummy variable denoting whether both posts used the same slur. For each LRM, log reasoning tokens were regressed on the same dummy variable (see Tables~S5 and S11 for the full specifications, including the pair-level models). The other conjoint attributes were not controlled for, since they were fully randomized and varied independently of the same-slur dummy. For single decisions, log RT was regressed on each model's log reasoning tokens (Table~S7). The coefficient on log tokens is the elasticity of RT with respect to CoT length. The comparison across quintiles (\autoref{fig:fig3}A) replaced log tokens with indicators for the quintiles of each model's reasoning tokens, so that each quintile was compared to the quintile with the fewest tokens (Table~S9). All models estimating RTs control for the log task number, which accounts for learning as subjects performed more tasks (Fig.~S1) and include subject fixed effects and standard errors clustered by subject. The percentage differences reported for the same-slur and quintile comparisons are calculated as $100 \times (\exp(\hat{\beta}) - 1)$, since a coefficient on a logged outcome approximates a percentage difference only when it is small and several of the estimates here are large. For robustness, we repeated the regressions (a) including the outlier observations and (b) using an alternative timing measure, the last click on each choice task. In each case, the results of these checks are consistent with the main specifications (Tables~S6, S8, and S10).

\subsection{Correlations}
For the aggregated correlations, we grouped pairs into cells defined by the combinations of variables in each pair of posts, and correlated the cell means of adjusted log RT and log reasoning tokens, using percentile bootstrap confidence intervals over cells (1{,}000 resamples). The human noise ceiling is the square root of the split-half reliability of the human cell means, estimated as in de Varda et al. \citep{de_varda_cost_2025}. For single decisions, which cannot be split because each pair was judged once, the regression benchmark is the square root of the share of variance in adjusted log RT explained by all manipulated attributes of both posts (adjusted $R^2$). The SI (``Correlations of averaged decisions''; Table~S3) describes the cells, the random grouping used in the placebo analysis, and the intervals.

\subsection{Content analyses of reasoning traces}
The qualitative results were obtained by coding a stratified random sample of reasoning traces from Gemini 2.5 Pro ($n=200$), consisting of 100 pairs where the same slur was used in both posts and 100 pairs where it varied. The reasoning traces are summaries of CoTs from Gemini (the ``raw'' CoTs were not provided). Each was coded by the author and two undergraduate research assistants to determine whether or not each of the factors manipulated in the experiment was mentioned. Each CoT was coded for mentions of slur use (including the generic insult), cursing, user race, user gender, post topic, like count (engagement), and replies. To ensure reliable coding, research assistants were given detailed instructions and were provided with feedback after coding ten practice examples. In the analyses presented here, a reasoning trace was considered to include a factor if two or more raters marked it as present. Chi-squared tests were used to assess whether the proportion of traces mentioning each factor differed across the two sets of posts.

\subsection{AI assistance}
Anthropic's Claude Code was used to help write code, analyze data, and edit the manuscript. Additional literature review was conducted using Claude Code and OpenAI's Deep Research feature in ChatGPT. The latest available versions of each model were used, primarily variants of Claude Opus 4 and 5 and GPT-5.

\section{Acknowledgments}
I thank Nicholas Armeli and Juliana Ospina for research assistance. This study was previously presented at the 2025 Summer Institute in Computational Social Science at the University of Pennsylvania, the 2025 American Sociological Association Annual Meeting, the Quantitative and Computational Methods Workshop in the Department of Sociology at the University of Notre Dame, the Social Dynamics Lab at Cornell University, and the 2026 International Conference for Computational Social Science at the University of Vermont. I thank participants at these events and the reviewers for their feedback. All errors are my own.%

\section{Competing interest}
No competing interests to declare.

\section{Funding}
This research was supported by a Foundational Integrity Research award from Meta and API credits provided by OpenAI's Researcher Access Program.

\section{Author contributions}
T.R.D.: Funding acquisition, Investigation, Conceptualization, Data curation, Formal analysis, Methodology, Software, Validation, Visualization, Writing---original draft, Writing---review \& editing.

\section{Data availability}
Data and code required to replicate the results will be made available in a public repository upon acceptance.

\section{Ethics statement}
This paper reports findings from a human subjects study. Informed consent was obtained from all human subjects, and the experiment was approved by the Rutgers University Institutional Review Board (\#Pro2023002017 \& \#Mod2024000438).

\bibliographystyle{oup-plain-nexus}
\bibliography{references}

\begin{thebibliography}{10}

\bibitem{anderson_adaptive_1990}
John~R. Anderson.
\newblock {\em The Adaptive Character of Thought}.
\newblock Lawrence Erlbaum Associates, Hillsdale, NJ, 1990.

\bibitem{arthur_thinking_2026}
Henry Arthur.
\newblock Thinking effort aligns between humans and reasoning models in abductive reasoning.
\newblock In {\em Findings of the {Association} for {Computational} {Linguistics}: {EMNLP} 2026}, pages 11136--11149. Association for Computational Linguistics, 2026.

\bibitem{bai_explicitly_2025}
Xuechunzi Bai, Angelina Wang, Ilia Sucholutsky, and Thomas~L. Griffiths.
\newblock Explicitly unbiased large language models still form biased associations.
\newblock {\em Proceedings of the National Academy of Sciences}, 122(8):e2416228122, 2025.

\bibitem{bao-etal-2025-likely}
Guangsheng Bao, Hongbo Zhang, Cunxiang Wang, Linyi Yang, and Yue Zhang.
\newblock How likely do {LLM}s with {C}o{T} mimic human reasoning?
\newblock In {\em Proceedings of the 31st International Conference on Computational Linguistics}, pages 7831--7850. Association for Computational Linguistics, 2025.

\bibitem{bettman_componential_1990}
James~R. Bettman, Eric~J. Johnson, and John~W. Payne.
\newblock A componential analysis of cognitive effort in choice.
\newblock {\em Organizational Behavior and Human Decision Processes}, 45(1):111--139, 1990.

\bibitem{bhatia_computational_2025}
Sudeep Bhatia, Simon~T. van Baal, Feiyi Wang, and Lukasz Walasek.
\newblock Computational analysis of 100 {K} choice dilemmas: {Decision} attributes, trade-off structures, and model-based prediction.
\newblock {\em Proceedings of the National Academy of Sciences}, 122(17):e2406489122, 2025.

\bibitem{binz_foundation_2025}
Marcel Binz et~al.
\newblock A foundation model to predict and capture human cognition.
\newblock {\em Nature}, 644(8078):1002--1009, 2025.

\bibitem{brodeur_performance_2026}
Peter~G. Brodeur et~al.
\newblock Performance of a large language model on the reasoning tasks of a physician.
\newblock {\em Science}, 392(6797):524--527, 2026.

\bibitem{chen_reasoning_2025}
Yanda Chen et~al.
\newblock Reasoning {Models} {Don}'t {Always} {Say} {What} {They} {Think}, 2025.
\newblock arXiv:2505.05410 [cs].

\bibitem{cheung_large_2025}
Vanessa Cheung, Maximilian Maier, and Falk Lieder.
\newblock Large language models show amplified cognitive biases in moral decision-making.
\newblock {\em Proceedings of the National Academy of Sciences}, 122(25):e2412015122, 2025.

\bibitem{davidson-multimodal-2025}
Thomas Davidson.
\newblock Multimodal large language models can make context-sensitive hate speech evaluations aligned with human judgement.
\newblock {\em Nature Human Behaviour}, 10(3):514--530, 2026.

\bibitem{davidson_racial_2019}
Thomas Davidson, Debasmita Bhattacharya, and Ingmar Weber.
\newblock Racial {Bias} in {Hate} {Speech} and {Abusive} {Language} {Detection} {Datasets}.
\newblock In {\em Proceedings of the {Third} {Workshop} on {Abusive} {Language} {Online}}, pages 25--35. ACL, 2019.

\bibitem{davidson_automated_2017}
Thomas Davidson, Dana Warmsley, Michael Macy, and Ingmar Weber.
\newblock Automated {Hate} {Speech} {Detection} and the {Problem} of {Offensive} {Language}.
\newblock In {\em Proceedings of the 11th {International} {Conference} on {Web} and {Social} {Media} ({ICWSM})}, pages 512--515, 2017.

\bibitem{de_varda_cost_2025}
Andrea~Gregor de~Varda, Ferdinando~Pio D’Elia, Hope Kean, Andrew Lampinen, and Evelina Fedorenko.
\newblock The cost of thinking is similar between large reasoning models and humans.
\newblock {\em Proceedings of the National Academy of Sciences}, 122(47):e2520077122, 2025.

\bibitem{de_varda_reply_2026-2}
Andrea~Gregor de~Varda, Ferdinando~Pio D’Elia, Hope Kean, Andrew Lampinen, and Evelina Fedorenko.
\newblock Reply to {Dujmović}: {The} alignment in cost between human and model reasoning is an empirical phenomenon worth explaining.
\newblock {\em Proceedings of the National Academy of Sciences}, 123(4):e2536153123, 2026.

\bibitem{de_varda_reply_2026}
Andrea~Gregor de~Varda, Ferdinando~Pio D’Elia, Hope Kean, Andrew Lampinen, and Evelina Fedorenko.
\newblock Reply to {Hu}: {Reasoning} traces need not be faithful to account for the cost of human thinking.
\newblock {\em Proceedings of the National Academy of Sciences}, 123(17):e2606498123, 2026.

\bibitem{de_varda_reply_2026-1}
Andrea~Gregor de~Varda, Ferdinando~Pio D’Elia, Hope Kean, Andrew Lampinen, and Evelina Fedorenko.
\newblock Reply to {Vankov} et al.: {Reasoning} traces are linked to accuracy and capture key dimensions of problem complexity.
\newblock {\em Proceedings of the National Academy of Sciences}, 123(12):e2603574123, 2026.

\bibitem{dixon_measuring_2018}
Lucas Dixon, John Li, Jeffrey Sorensen, Nithum Thain, and Lucy Vasserman.
\newblock Measuring and {Mitigating} {Unintended} {Bias} in {Text} {Classification}.
\newblock In {\em Proceedings of the 2018 {Conference} on {AI}, {Ethics}, and {Society}}, pages 67--73. ACM Press, 2018.

\bibitem{dujmovic_no_2026}
Marin Dujmović.
\newblock No deep insights into the alignment between human and deep learning reasoning processes: {Thoughts} on de {Varda} et al. (2025).
\newblock {\em Proceedings of the National Academy of Sciences}, 123(4):e2533685123, 2026.

\bibitem{gigerenzer_reasoning_1996}
Gerd Gigerenzer and Daniel~G. Goldstein.
\newblock Reasoning the fast and frugal way: {Models} of bounded rationality.
\newblock {\em Psychological Review}, 103(4):650--669, 1996.

\bibitem{goodkind_predictive_2018}
Adam Goodkind and Klinton Bicknell.
\newblock Predictive power of word surprisal for reading times is a linear function of language model quality.
\newblock In {\em Proceedings of the 8th {Workshop} on {Cognitive} {Modeling} and {Computational} {Linguistics} ({CMCL} 2018)}, pages 10--18. Association for Computational Linguistics, 2018.

\bibitem{greenwald_measuring_1998}
Anthony Greenwald, Debbie McGhee, and Jordan Schwartz.
\newblock Measuring {Individual} {Differences} in {Implicit} {Cognition}.
\newblock {\em Journal of Personality and Social Psychology}, 74(6):1464--1480, 1998.

\bibitem{guan_deliberative_2025}
Melody~Y. Guan et~al.
\newblock Deliberative {Alignment}: {Reasoning} {Enables} {Safer} {Language} {Models}, 2025.
\newblock arXiv:2412.16339 [cs].

\bibitem{guo_deepseek-r1_2025}
Daya Guo et~al.
\newblock {DeepSeek}-{R1} incentivizes reasoning in {LLMs} through reinforcement learning.
\newblock {\em Nature}, 645(8081):633--638, 2025.

\bibitem{hainmueller_validating_2015}
Jens Hainmueller, Dominik Hangartner, and Teppei Yamamoto.
\newblock Validating vignette and conjoint survey experiments against real-world behavior.
\newblock {\em Proceedings of the National Academy of Sciences}, 112(8):2395--2400, 2015.

\bibitem{hainmueller_causal_2014}
Jens Hainmueller, Daniel~J. Hopkins, and Teppei Yamamoto.
\newblock Causal {Inference} in {Conjoint} {Analysis}: {Understanding} {Multidimensional} {Choices} via {Stated} {Preference} {Experiments}.
\newblock {\em Political Analysis}, 22(1):1--30, 2014.

\bibitem{hartigan_dip_1985}
J.~A. Hartigan and P.~M. Hartigan.
\newblock The dip test of unimodality.
\newblock {\em The Annals of Statistics}, 13(1):70--84, 1985.

\bibitem{hartvigsen_toxigen_2022}
Thomas Hartvigsen, Saadia Gabriel, Hamid Palangi, Maarten Sap, Dipankar Ray, and Ece Kamar.
\newblock {ToxiGen}: {A} {Large}-{Scale} {Machine}-{Generated} {Dataset} for {Adversarial} and {Implicit} {Hate} {Speech} {Detection}.
\newblock In {\em Proceedings of the 60th {Annual} {Meeting} of the {Association} for {Computational} {Linguistics}}, pages 3309--3326. ACL, 2022.

\bibitem{hick_rate_1952}
W.~E. Hick.
\newblock On the {Rate} of {Gain} of {Information}.
\newblock {\em Quarterly Journal of Experimental Psychology}, 4(1):11--26, 1952.

\bibitem{hofmann_ai_2024}
Valentin Hofmann, Pratyusha~Ria Kalluri, Dan Jurafsky, and Sharese King.
\newblock {AI} generates covertly racist decisions about people based on their dialect.
\newblock {\em Nature}, 633(8028):147--154, 2024.

\bibitem{hu_thinking_2026}
Yueqing Hu.
\newblock “{Thinking} traces” in large reasoning models: {Cognitive} cost or performative scaffolding?
\newblock {\em Proceedings of the National Academy of Sciences}, 123(17):e2604554123, 2026.

\bibitem{kojima_large_2022}
Takeshi Kojima, Shixiang~Shane Gu, Machel Reid, Yutaka Matsuo, and Yusuke Iwasawa.
\newblock Large language models are zero-shot reasoners.
\newblock In {\em Proceedings of the 36th {International} {Conference} on {Neural} {Information} {Processing} {Systems}}, pages 22199--22213. Curran Associates Inc., 2022.

\bibitem{korbak_chain_2025}
Tomek Korbak et~al.
\newblock Chain of {Thought} {Monitorability}: {A} {New} and {Fragile} {Opportunity} for {AI} {Safety}, 2025.
\newblock arXiv:2507.11473 [cs.AI].

\bibitem{krosnick_response_1991}
Jon~A. Krosnick.
\newblock Response strategies for coping with the cognitive demands of attitude measures in surveys.
\newblock {\em Applied Cognitive Psychology}, 5(3):213--236, 1991.

\bibitem{kwon_efficient_2023}
Woosuk Kwon, Zhuohan Li, Siyuan Zhuang, Ying Sheng, Lianmin Zheng, Cody~Hao Yu, Joseph Gonzalez, Hao Zhang, and Ion Stoica.
\newblock Efficient {Memory} {Management} for {Large} {Language} {Model} {Serving} with {PagedAttention}.
\newblock In {\em Proceedings of the 29th {Symposium} on {Operating} {Systems} {Principles}}, pages 611--626. Association for Computing Machinery, 2023.

\bibitem{lampinen_language_2024}
Andrew~K Lampinen, Ishita Dasgupta, Stephanie C~Y Chan, Hannah~R Sheahan, Antonia Creswell, Dharshan Kumaran, James~L McClelland, and Felix Hill.
\newblock Language models, like humans, show content effects on reasoning tasks.
\newblock {\em PNAS Nexus}, 3(7):pgae233, 2024.

\bibitem{lanham_measuring_2023}
Tamera Lanham et~al.
\newblock Measuring {Faithfulness} in {Chain}-of-{Thought} {Reasoning}, 2023.
\newblock arXiv:2307.13702 [cs].

\bibitem{lee_implicit-bias-like_2026}
Messi H.~J. Lee and Calvin~K. Lai.
\newblock Implicit-bias-like patterns in reasoning models.
\newblock {\em Nature Machine Intelligence}, 8(9):1457--1465, 2026.

\bibitem{lightman_lets_2023}
Hunter Lightman, Vineet Kosaraju, Yuri Burda, Harrison Edwards, Bowen Baker, Teddy Lee, Jan Leike, John Schulman, Ilya Sutskever, and Karl Cobbe.
\newblock Let's {Verify} {Step} by {Step}.
\newblock In {\em The Twelfth International Conference on Learning Representations}, 2023.

\bibitem{loru_simulation_2025}
Edoardo Loru, Jacopo Nudo, Niccolò Di~Marco, Alessandro Santirocchi, Roberto Atzeni, Matteo Cinelli, Vincenzo Cestari, Clelia Rossi-Arnaud, and Walter Quattrociocchi.
\newblock The simulation of judgment in {LLMs}.
\newblock {\em Proceedings of the National Academy of Sciences}, 122(42):e2518443122, 2025.

\bibitem{muennighoff_s1_2025}
Niklas Muennighoff, Zitong Yang, Weijia Shi, Xiang~Lisa Li, Li~Fei-Fei, Hannaneh Hajishirzi, Luke Zettlemoyer, Percy Liang, Emmanuel Candès, and Tatsunori Hashimoto.
\newblock s1: {Simple} test-time scaling.
\newblock In {\em Proceedings of the 2025 {Conference} on {Empirical} {Methods} in {Natural} {Language} {Processing}}, pages 20275--20321. Association for Computational Linguistics, 2025.

\bibitem{mulligan_response_2003}
Kenneth Mulligan, J.~Tobin Grant, Stephen~T. Mockabee, and Joseph~Quin Monson.
\newblock Response {Latency} {Methodology} for {Survey} {Research}: {Measurement} and {Modeling} {Strategies}.
\newblock {\em Political Analysis}, 11(3):289--301, 2003.

\bibitem{oh_transformer-based_2023}
Byung-Doh Oh and William Schuler.
\newblock Transformer-{Based} {Language} {Model} {Surprisal} {Predicts} {Human} {Reading} {Times} {Best} with {About} {Two} {Billion} {Training} {Tokens}.
\newblock In {\em Findings of the {Association} for {Computational} {Linguistics}: {EMNLP} 2023}, pages 1915--1921. Association for Computational Linguistics, 2023.

\bibitem{openai_technical_2025}
OpenAI.
\newblock Technical {Report}: {Performance} and baseline evaluations of gpt-oss-safeguard-120b and gpt-oss-safeguard-20b.
\newblock Technical report, OpenAI, 2025.

\bibitem{openai_openai_2024}
OpenAI et~al.
\newblock {OpenAI} o1 {System} {Card}, 2024.
\newblock arXiv:2412.16720 [cs].

\bibitem{payne_adaptive_1993}
John~W. Payne, James~R. Bettman, and Eric~J. Johnson.
\newblock {\em The Adaptive Decision Maker}.
\newblock Cambridge University Press, Cambridge, 1993.

\bibitem{schneider_hicks_2018}
Robert~W. Proctor and Darryl~W. Schneider.
\newblock Hick's law for choice reaction time: {A} review.
\newblock {\em Quarterly Journal of Experimental Psychology}, 71(6):1281--1299, 2018.

\bibitem{rottger_hatecheck_2021}
Paul Röttger, Bertie Vidgen, Dong Nguyen, Zeerak Waseem, Helen Margetts, and Janet Pierrehumbert.
\newblock {HateCheck}: {Functional} {Tests} for {Hate} {Speech} {Detection} {Models}.
\newblock In {\em Proceedings of the 59th {Annual} {Meeting} of the {Association} for {Computational} {Linguistics} and the 11th {International} {Joint} {Conference} on {Natural} {Language} {Processing}}, pages 41--58. ACL, 2021.

\bibitem{sap_risk_2019}
Maarten Sap, Dallas Card, Saadia Gabriel, Yejin Choi, and Noah~A Smith.
\newblock The {Risk} of {Racial} {Bias} in {Hate} {Speech} {Detection}.
\newblock In {\em Proceedings of the 57th {Annual} {Meeting} of the {Association} for {Computational} {Linguistics}}, pages 1668--1678. ACL, 2019.

\bibitem{illusion-of-thinking}
Parshin Shojaee, Iman Mirzadeh, Keivan Alizadeh, Maxwell Horton, Samy Bengio, and Mehrdad Farajtabar.
\newblock The illusion of thinking: Understanding the strengths and limitations of reasoning models via the lens of problem complexity.
\newblock In {\em Advances in {Neural} {Information} {Processing} {Systems}}, 2025.

\bibitem{simon_rational_1956}
Herbert~A. Simon.
\newblock Rational choice and the structure of the environment.
\newblock {\em Psychological Review}, 63(2):129--138, 1956.

\bibitem{sloman_empirical_1996}
Steven~A. Sloman.
\newblock The empirical case for two systems of reasoning.
\newblock {\em Psychological Bulletin}, 119(1):3--22, 1996.

\bibitem{snell_scaling_2024}
Charlie Snell, Jaehoon Lee, Kelvin Xu, and Aviral Kumar.
\newblock Scaling {LLM} {Test}-{Time} {Compute} {Optimally} can be {More} {Effective} than {Scaling} {Model} {Parameters}, 2024.
\newblock arXiv:2408.03314 [cs].

\bibitem{stanovich_individual_2000}
Keith~E. Stanovich and Richard~F. West.
\newblock Individual differences in reasoning: {Implications} for the rationality debate?
\newblock {\em Behavioral and Brain Sciences}, 23(5):645--665, 2000.

\bibitem{team_kimi-vl_2025}
Kimi Team et~al.
\newblock Kimi-{VL} {Technical} {Report}, 2025.
\newblock arXiv:2504.07491 [cs.CV].

\bibitem{turpin_language_2023}
Miles Turpin, Julian Michael, Ethan Perez, and Samuel Bowman.
\newblock Language {Models} {Don}'t {Always} {Say} {What} {They} {Think}: {Unfaithful} {Explanations} in {Chain}-of-{Thought} {Prompting}.
\newblock {\em Advances in Neural Information Processing Systems}, 36:74952--74965, 2023.

\bibitem{tversky_judgment_1974}
Amos Tversky and Daniel Kahneman.
\newblock Judgment under {Uncertainty}: {Heuristics} and {Biases}.
\newblock {\em Science}, 185(4157):1124--1131, 1974.

\bibitem{vankov_correlations_2026}
Ivan~I. Vankov, Federico Adolfi, Rachel~F. Heaton, Guillermo Puebla, and Jeffrey~S. Bowers.
\newblock Correlations without causation do not support claims of human–{LLM} reasoning alignment.
\newblock {\em Proceedings of the National Academy of Sciences}, 123(12):e2536362123, 2026.

\bibitem{wei_chain--thought_2022}
Jason Wei, Xuezhi Wang, Dale Schuurmans, Maarten Bosma, Brian Ichter, Fei Xia, Ed~Chi, Quoc~V Le, and Denny Zhou.
\newblock Chain-of-{Thought} {Prompting} {Elicits} {Reasoning} in {Large} {Language} {Models}.
\newblock In {\em Advances in {Neural} {Information} {Processing} {Systems}}, volume~35, pages 24824--24837. Curran Associates, Inc., 2022.

\bibitem{yang_qwen3_2025}
An~Yang et~al.
\newblock Qwen3 {Technical} {Report}, 2025.
\newblock arXiv:2505.09388 [cs.CL].

\bibitem{yao_tree_2023}
Shunyu Yao, Dian Yu, Jeffrey Zhao, Izhak Shafran, Thomas~L. Griffiths, Yuan Cao, and Karthik Narasimhan.
\newblock Tree of thoughts: deliberate problem solving with large language models.
\newblock In {\em Proceedings of the 37th {International} {Conference} on {Neural} {Information} {Processing} {Systems}}, pages 11809--11822. Curran Associates Inc., 2023.

\bibitem{yue_does_2026}
Yang Yue, Zhiqi Chen, Rui Lu, Andrew Zhao, Zhaokai Wang, Yang Yue, Shiji Song, and Gao Huang.
\newblock Does {Reinforcement} {Learning} {Really} {Incentivize} {Reasoning} {Capacity} in {LLMs} {Beyond} the {Base} {Model}?
\newblock {\em Advances in Neural Information Processing Systems}, 38:57654--57689, 2026.

\bibitem{zhang_speeding_2014}
Chan Zhang and Frederick Conrad.
\newblock Speeding in {Web} {Surveys}: {The} tendency to answer very fast and its association with straightlining.
\newblock {\em Survey Research Methods}, 8(2):127--135, 2014.

\bibitem{zhang_biasx_2023}
Yiming Zhang, Sravani Nanduri, Liwei Jiang, Tongshuang Wu, and Maarten Sap.
\newblock {BiasX}: “{Thinking} {Slow}” in {Toxic} {Content} {Moderation} with {Explanations} of {Implied} {Social} {Biases}.
\newblock In {\em Proceedings of the 2023 {Conference} on {Empirical} {Methods} in {Natural} {Language} {Processing}}, pages 4920--4932. Association for Computational Linguistics, 2023.

\bibitem{zhou_cobra_2023}
Xuhui Zhou, Hao Zhu, Akhila Yerukola, Thomas Davidson, Jena~D. Hwang, Swabha Swayamdipta, and Maarten Sap.
\newblock {COBRA} {Frames}: {Contextual} {Reasoning} about {Effects} and {Harms} of {Offensive} {Statements}.
\newblock In {\em Findings of the {Association} for {Computational} {Linguistics}: {ACL} 2023}, pages 6294--6315. Association for Computational Linguistics, 2023.

\end{thebibliography}

\clearpage
\raggedbottom
\renewcommand{\theHtable}{S\arabic{table}}
\renewcommand{\theHfigure}{S\arabic{figure}}
\section*{Supporting Information}
\setcounter{table}{0}
\renewcommand{\thetable}{S\arabic{table}}
\setcounter{figure}{0}
\renewcommand{\thefigure}{S\arabic{figure}}

\subsection*{Conjoint design and human subjects experiment}

This study used data from a conjoint experiment on content moderation originally reported in an earlier study \citep{davidson-multimodal-2025}. Conjoint experiments allow many attributes to be manipulated simultaneously \citep{hainmueller_causal_2014}. This design facilitated the analysis of how multiple linguistic and contextual factors influence content moderation decisions. Building on work showing how synthetic media can augment hate speech datasets and how template designs can audit AI content moderation systems \citep{dixon_measuring_2018,rottger_hatecheck_2021,hartvigsen_toxigen_2022}, the experiment used synthetic social media posts (resembling X/Twitter posts) that were constructed to depict speech that may or may not warrant moderation under a hate speech policy. The text of the posts was generated with GPT-4 and edited by hand to ensure consistency. Each post varied across several dimensions (see \autoref{table:conjoint}), and posts with all combinations of attributes and values were created, yielding 210{,}000 distinct posts. The study used a forced-choice design, which is standard in conjoint methodology and performs favorably compared to alternative approaches \citep{hainmueller_validating_2015}. Each task required a choice between two options, in this case, selecting which of two posts should be prioritized for moderation.

\begin{table}[!htbp]
    \centering
    \caption{\textbf{Attributes varied in synthetic social media posts}}
\label{table:conjoint}
    \begin{tabular}{lll}
        \toprule
        \textbf{Attribute} & \textbf{Values} \\
        \midrule
        Slur & None, Generic insult, Sexism, Homophobia, Racism, \\
        & Reclaimed slur, Anti-White \\
        \addlinespace
        Cursing & No curse, Curse \\
        \addlinespace
        Topic\textsuperscript{\textdagger} & Sports, Politics, Entertainment, Workplace, Everyday \\
        Identity\textsuperscript{\textdagger} & Black female, Black male, White female, White male, Anonymous \\
        \addlinespace
        Reply & None, Agree, Disagree \\
        \addlinespace
        Engagement & High [25--50 likes], Low [0--5 likes] \\
        \bottomrule
        \small{\textsuperscript{\textdagger} Ten variations for each category.} & & \\
        
    \end{tabular}
\end{table}

To assess how closely the synthetic posts resemble real social media content, we compared the 700 distinct post texts from 50 templates with 24{,}783 tweets from a widely used corpus of hate speech \citep{davidson_automated_2017}. \autoref{tab:realism} compares different measures of length, complexity, and readability. The median post is somewhat longer than the median tweet (15 words versus 12) and uses longer words (4.8 letters per word versus 4.0, and 23\% of words longer than six letters versus 10\%), and its Flesch--Kincaid reading grade is slightly higher (4.5 versus 3.7). The interquartile ranges of the posts and the tweets overlap on every measure. The posts were intended to be of similar length, and each template was written to accommodate any slur while holding everything else constant, so there is less variability than among real tweets. Taken together, some departures from real tweets are expected, since the design prioritizes internal and construct validity, but the posts were designed to be realistic enough for human subjects and large reasoning models (LRMs) to evaluate them as they would genuine user-generated content.

\begin{table}[!htbp]
    \addedtable
    \centering
    \caption{Comparison of synthetic posts and real tweets.}
    \label{tab:realism}
    \begin{threeparttable}
    \begin{tabular}{lcc}
    \toprule
     & Synthetic posts & Real tweets \\
    \midrule
    Words per post & 15.0 (14.0--17.0) & 12.0 (7.0--18.0) \\
    Letters per word & 4.8 (4.4--5.3) & 4.0 (3.6--4.5) \\
    Long words (\%) & 23.1 (14.3--33.3) & 10.0 (0.0--17.6) \\
    Reading grade & 4.5 (3.0--6.6) & 3.7 (1.3--5.8) \\
    \midrule
    $N$ & 700 & 24{,}783 \\
    \bottomrule
    \end{tabular}
    \begin{tablenotes}[flushleft]
    \footnotesize
    \item[] \textit{Note}: Medians, with interquartile ranges in parentheses. Links, mentions, and emojis were removed before measuring. Reading grade was not computed for the 645 tweets left with fewer than three words, and letters per word and long words exclude texts left with none after preprocessing.
    \end{tablenotes}
    \end{threeparttable}
\end{table}

Subjects were recruited via Prolific, an online opt-in panel provider. The inclusion criteria required subjects to be adults aged 18 to 65 residing in the United States who spoke English. To ensure familiarity with social media, subjects were included if they had reported using one or more of the following social media platforms: Facebook, Reddit, Twitter, YouTube, TikTok, or Instagram. The sample was stratified by self-reported ethnicity, sex, sexual orientation, and political party affiliation. The quotas were based on US Census statistics but were set to oversample Black and LGBT+ subjects to ensure sufficient representation of groups often targeted by hate speech. Additionally, the study was restricted to people who had completed at least 50 studies on Prolific with a 99\% approval rate to help guarantee data quality. Prolific requires that participants opt into studies involving sensitive content. Subjects were also required to use a desktop computer to ensure that the conjoint images were rendered legibly, and any subject who attempted to take part using a screen with a resolution lower than that of most desktops or tablets ($<1280\times720$ pixels) was removed and replaced with another eligible subject. Additional details on the design of the posts and the human subjects experiment are reported in related work \cite{davidson-multimodal-2025}. 

\subsection*{Conjoint task}

The conjoint task consisted of a forced-choice question in which human subjects or LRMs had to select one post from a pair. In each case, a policy was provided, and the instructions stated that the post that was more likely to violate the policy should be selected. We used the policy adopted by X (\url{https://help.x.com/en/rules-and-policies/hateful-conduct-policy}), which reads: 

\begin{quote}
\textit{We are committed to combating abuse motivated by hatred, prejudice or intolerance, particularly abuse that seeks to silence the voices of those who have been historically marginalized. For this reason, we prohibit behavior that targets individuals or groups with abuse based on their perceived membership in a protected category. You may not directly attack other people on the basis of race, ethnicity, national origin, caste, sexual orientation, gender, gender identity, religious affiliation, age, disability, or serious disease.}
\end{quote}

This policy was selected because it is relatively short and easy to read, it identifies key targets of hate speech manipulated in the experiment, and it matches the social media context of the task.

\subsection*{Response time and task order}

\begin{figure}[!htbp]
\centering

\includegraphics[width=\textwidth]{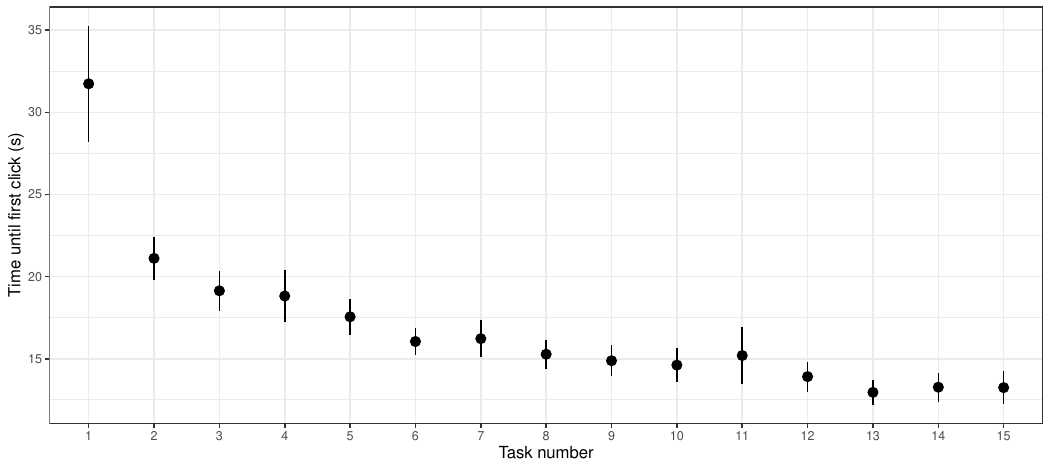}
\caption{First click by task number. The figure shows the mean time to first click across all subjects ($N = 1{,}854$) by task number, with 95\% confidence intervals.}
\label{fig:firstclick}
\end{figure}

The data analyzed in this study consist of the response times (RTs) on each conjoint task. Each subject rated 15 distinct pairs of posts. The LRM experiments, described in the following section, used the same set of pairs. \autoref{fig:firstclick} shows the average RT across subjects for each of the 15 tasks. Subjects took significantly longer on the first pair, taking 31.7 seconds on average to select an option. The second pair was faster, taking 21.1 seconds on average, and RTs continued to decrease as subjects completed more tasks. The average time to decide on the final pair was 13.2 seconds. This pattern shows that subjects responded more quickly as they became more familiar with the task. To account for this learning process, the logarithm of the task number was used, along with subject fixed effects, to construct the adjusted RT, and was included as a control in the regression analyses, which use the unadjusted RTs.

\subsection*{Prompt and LRM experiments}

The following prompt was shown to each model, along with each pair of posts:

\begin{quote}
You are a content moderation expert. You will receive pairs of images, A and B. Your task is to decide which image should be prioritized for manual review according to the following policy: [POLICY] The output should always be one of the two options: A or B. Only output a single character with no additional tokens. Never refuse to answer.
\end{quote}

In the scripts for Gemini, Grok, and the open-weights models, the second sentence read ``You will receive pairs of images labeled as `Image A' and `Image B'\,'' due to a difference in the interface. The implementation of the experiments varied across the closed-weights and open-weights models.

\subsection*{Closed-weights models}

All three closed-weights frontier reasoning models were accessed using proprietary APIs. We used similar configurations for each model to minimize variation due to parameterization. o3 and Gemini 2.5 Pro had parameters controlling the reasoning effort. To avoid arbitrarily limiting the models' reasoning capacity, o3 was set to the highest level of reasoning effort, and Gemini 2.5 Pro used a dynamic thinking budget, which let the model decide how much to reason. Grok 4 did not offer this option. Gemini 2.5 Pro and Grok 4 also accepted temperature parameters, which can control the amount of stochastic variation in the results. Both were set to zero to ensure the most likely token was returned. This option was not available for o3. Gemini 2.5 Pro was the only closed-weights model to provide the reasoning traces at the time of the analysis. The models were priced based on the number of tokens in the input (including text and images), the output, and the reasoning token usage. Each closed-weights experiment cost between \$350 and \$700.

o3 reported its reasoning tokens in blocks of 64, so its counts took few distinct values. For Fig.~3 of the main text, each of o3's counts was replaced by a value drawn uniformly from within 31.5 tokens on either side of it, and the dip test was computed on the logarithm of these values. All other analyses used the reported counts.

\subsection*{Open-weights models}
The three open-weights models were served using the vLLM inference framework \citep{kwon_efficient_2023} running on the Amarel high-performance computing cluster at Rutgers University, using a node equipped with four NVIDIA L40S GPUs. Each model received the same prompt as the closed-weights models, and the client script issued requests through the vLLM server's API. The server reported the total number of tokens generated, so a small, fixed number of tokens generated after the reasoning (the end-of-thinking marker, the one-letter answer, and the end-of-turn token) was removed before calculating the chain-of-thought (CoT) length. Each model was configured with the recommended sampling parameters from its documentation: Kimi-VL Thinking at temperature 0.8 and top-$p$ 1.0; Qwen 3.5 at temperature 1.0, top-$p$ 0.95, and top-$k$ 20; and Gemma 4 at temperature 1.0, top-$p$ 0.95, and top-$k$ 64.

Two parameters governed how much reasoning each model was permitted to perform. The server-side \texttt{max-model-len} is the maximum number of tokens the server keeps in its key-value cache for a single request, including the full prompt and all generated output. The client-side \texttt{max\_tokens} caps the number of tokens the model can emit for a given call---effectively the reasoning ``budget'' for the CoT plus the final answer. The recommended and maximum values varied across models, and setting values too high could result in out-of-memory errors. We thus experimented with each model to identify values that yielded consistent results with minimal errors. The baseline budgets (\texttt{max-model-len}/\texttt{max\_tokens}) were 8{,}192/4{,}096 for Kimi, 32{,}768/16{,}384 for Qwen, and 16{,}384/4{,}096 for Gemma. A small fraction of pairs had no valid answer after the main runs, mostly due to timeouts or token limits. Retry passes with larger token budgets and longer timeouts recovered results for all of these pairs.

\subsection*{Pairwise correlations}

\begin{figure}[p]
\centering
\begin{minipage}{0.84\textwidth}
\textbf{A}\par\vspace{2pt}
\includegraphics[width=\linewidth]{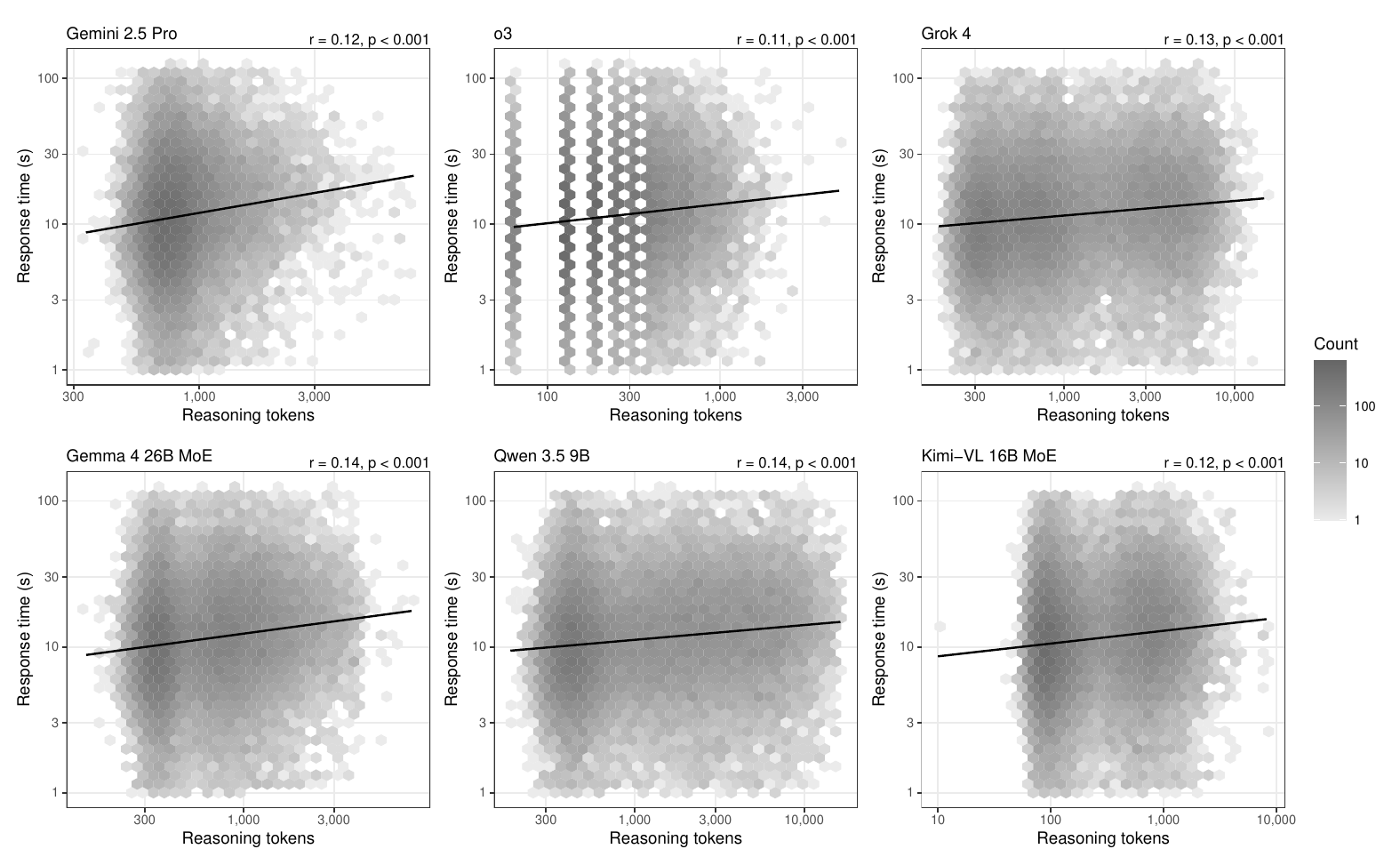}
\end{minipage}

\vspace{1em}
\begin{minipage}{0.84\textwidth}
\textbf{B}\par\vspace{2pt}
\includegraphics[width=\linewidth]{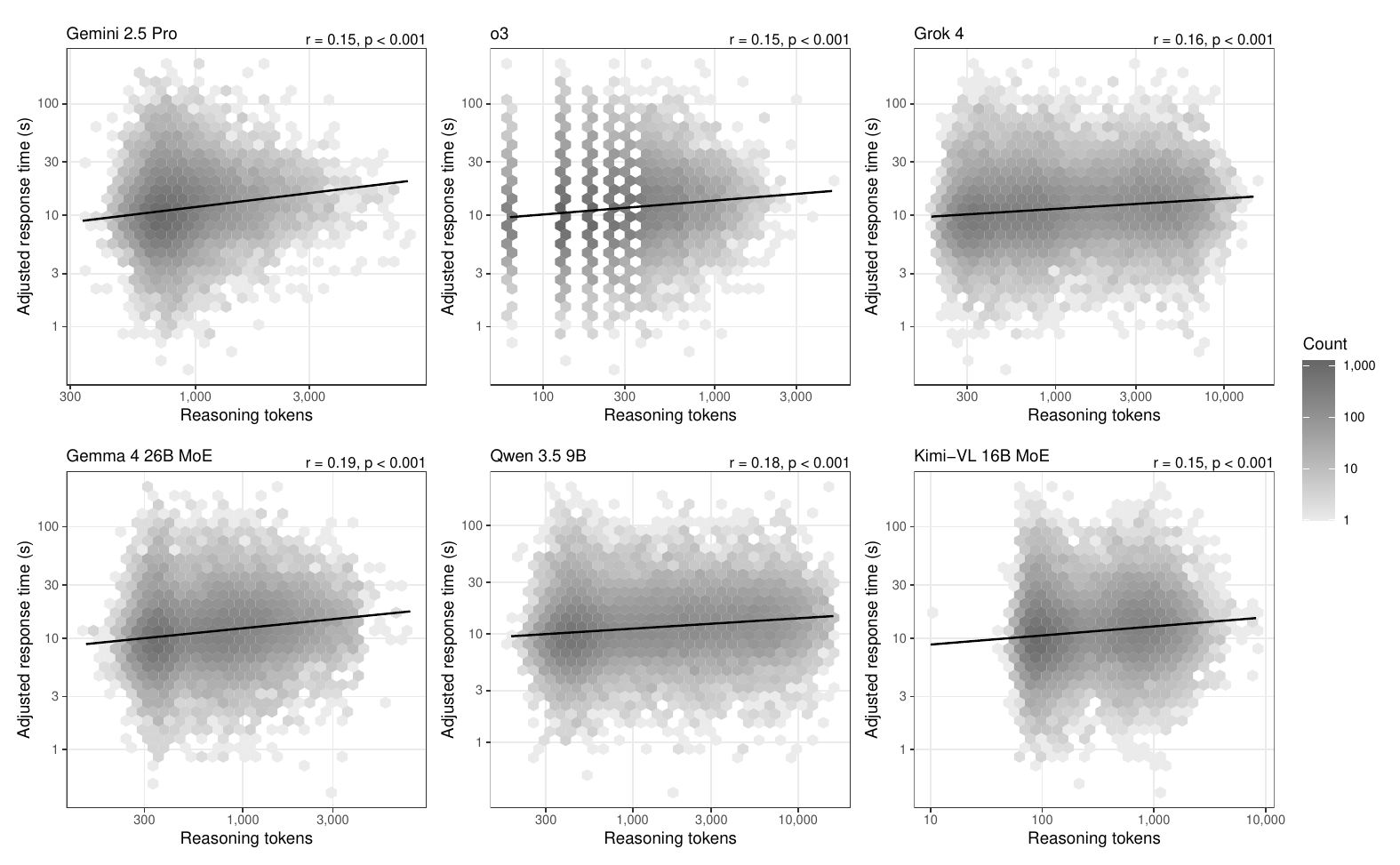}
\end{minipage}
\caption{Pairwise relationship between reasoning tokens and human RT for each LRM, on log-log axes. (A) Unadjusted RT. (B) RT adjusted for subject fixed effects and the log task number (see Materials and methods in the main text). Each panel is a hexagonal binning of the 27{,}071 analytic-sample pairs (27{,}052 for o3). The black line is a least-squares linear fit on the log-log scale, and the Pearson correlation between log reasoning tokens and log RT is reported in the upper right of each panel, alongside its significance level. Closed-weights models are in the top row (Gemini 2.5 Pro, o3, Grok 4); open-weights models are in the bottom row (Gemma 4 26B MoE, Qwen 3.5 9B, Kimi-VL 16B MoE).}\label{fig:scatters}
\end{figure}

\autoref{fig:scatters}A shows the pairwise Pearson correlation between reasoning tokens and human RTs for each LRM on log-log axes, using the analytic sample. The correlation is positive and statistically significant for all models. \autoref{fig:scatters}B shows the same correlations after adjusting for subject fixed effects and the log task number. The adjusted correlations are on average roughly 30\% higher than the corresponding raw correlations in \autoref{fig:scatters}A, indicating that some of the signal was suppressed by between-subject variation and the learning curve.

\subsection*{Correlations of averaged decisions}

Panel A of \autoref{tab:aggregation} reports the correlation between the cell means of adjusted log RT and log reasoning tokens at each level of aggregation shown in Fig.~4B of the main text. Each cell contains the pairs with the same two slurs (slur combination), the same two slurs and topics (slur $\times$ topic), or the same two slurs, topics, and user identities (slur $\times$ topic $\times$ identity), regardless of which post contained which. Cells with fewer than three pairs were dropped, which leaves 4{,}284 of the 6{,}300 possible cells at the finest level. The split-half reliability of the human cell means is computed as in de Varda et al. \citep{de_varda_cost_2025}: in each of 100 random splits, the pairs in each cell are divided into two halves, the half means are correlated across cells, and the Spearman--Brown correction is applied. The table reports the mean over these splits (the standard deviation is at most 0.02). For single decisions, the main effects of all manipulated attributes of both posts explain 6.6\% of the variance in adjusted log RT (adjusted $R^2$), so a predictor that captured their effects would correlate with single decisions at about $r = 0.26$. This regression benchmark, shown in Fig.~4B of the main text, is not a noise ceiling, since the regression includes every attribute but not the interactions among them, and each pair was judged only once. For averaged pairs, the noise ceiling in the table is the square root of the split-half reliability, with 95\% percentile intervals from 1{,}000 bootstrap resamples of cells.

Panel B of \autoref{tab:aggregation} reports a placebo analysis, which tests whether averaging alone raises the correlations. At each level, all pairs were assigned at random to as many groups of equal size as there are cells, so that the groups do not depend on the manipulated attributes (at the finest level the groups average 6.3 pairs, compared with 5.7 in Panel A, which leaves out the cells with fewer than three pairs), and the group means were correlated. The panel reports the mean and standard deviation of these correlations over 1{,}000 random groupings. The correlation stays close to its single-pair value (mean $r = 0.14$--$0.19$ across levels and models). None of the random groupings reached the observed correlation for any model at any level.

\begin{table}[!htbp]
    \addedtable
    \centering
    \scriptsize
    \setlength{\tabcolsep}{2.8pt}
    \caption{Correlation of averaged decisions.}
    \label{tab:aggregation}
    \begin{tabular}{lrrrccccccc}
    \toprule
     &  & Pairs & Split-half & Noise & Gemini &  &  &  &  &  \\
    Level & Cells & per cell & $\rho$ & ceiling & 2.5 Pro & o3 & Grok 4 & Gemma 4 & Qwen 3.5 & Kimi-VL \\
    \midrule
    \multicolumn{11}{l}{\textit{Panel A: Pairs grouped by design attributes}} \\
    \midrule
    Single pair & 27{,}071 & 1 & -- & -- & 0.15 & 0.15 & 0.16 & 0.19 & 0.18 & 0.15 \\
    Slur $\times$ topic $\times$ identity & 4{,}284 & 5.7 & 0.18 & 0.43 [0.38, 0.48] & 0.36 & 0.35 & 0.36 & 0.41 & 0.39 & 0.37 \\
    Slur $\times$ topic & 420 & 64 & 0.78 & 0.88 [0.86, 0.90] & 0.77 & 0.74 & 0.76 & 0.80 & 0.80 & 0.78 \\
    Slur combination & 28 & 967 & 0.98 & 0.99 [0.99, 0.99] & 0.88 & 0.85 & 0.86 & 0.93 & 0.91 & 0.90 \\
    \midrule[\heavyrulewidth]
    \multicolumn{11}{l}{\textit{Panel B: Placebo tests, means and standard deviations}} \\
    \midrule
    Slur $\times$ topic $\times$ identity & 4{,}284 & 6.3 &  &  & 0.15 & 0.15 & 0.16 & 0.19 & 0.18 & 0.15 \\
     &  &  &  &  & (0.01) & (0.01) & (0.01) & (0.01) & (0.01) & (0.01) \\
    Slur $\times$ topic & 420 & 64 &  &  & 0.15 & 0.15 & 0.17 & 0.19 & 0.18 & 0.15 \\
     &  &  &  &  & (0.05) & (0.05) & (0.05) & (0.05) & (0.05) & (0.05) \\
    Slur combination & 28 & 967 &  &  & 0.14 & 0.14 & 0.16 & 0.18 & 0.17 & 0.15 \\
     &  &  &  &  & (0.18) & (0.19) & (0.19) & (0.19) & (0.18) & (0.19) \\
    \bottomrule
    \end{tabular}
\end{table}

\subsubsection*{Representational similarity analysis}

\autoref{tab:similarity} compares the human and model matrices across the 21 combinations of two different slurs (Fig.~2B of the main text), as in a representational similarity analysis, with 95\% bootstrap confidence intervals over combinations and permutation $p$-values from 10{,}000 shuffles of the slur labels. The correlations across all 28 combinations are those in the slur-combination row of \autoref{tab:aggregation}, Panel A. The results are similar when the means are replaced with cell medians of the same adjusted log RTs and log reasoning tokens ($r = 0.83$--$0.93$ across all 28 combinations and $0.72$--$0.89$ across the 21 combinations of two different slurs).

\begin{table}[!htbp]
    \addedtable
    \centering
    \footnotesize
    \caption{Similarity of the human and model effort matrices across the 21 combinations of two different slurs.}
    \label{tab:similarity}
    \begin{tabular}{lcc}
    \toprule
    Model & $r$ [95\% CI] & Permutation $p$ \\
    \midrule
    Gemini 2.5 Pro & 0.72 [0.53, 0.88] & 0.001 \\
    o3 & 0.81 [0.69, 0.92] & 0.002 \\
    Grok 4 & 0.78 [0.62, 0.93] & 0.003 \\
    Gemma 4 26B MoE & 0.88 [0.77, 0.95] & $<$0.001 \\
    Qwen 3.5 9B & 0.86 [0.75, 0.93] & 0.001 \\
    Kimi-VL 16B MoE & 0.86 [0.73, 0.94] & 0.001 \\
    \bottomrule
    \end{tabular}
\end{table}

\subsection*{Regression tables}

The following tables show the estimates reported in the main text and the robustness checks, in the order in which the analyses appear in the main text. Each set of robustness checks is shown in two panels: Panel A replicates the main specification with outliers retained, and Panel B uses the log of the last-click time as the RT outcome (outliers removed). The RT models with outliers retained use 27{,}724 of the 27{,}810 decisions (27{,}705 for o3) since 84 decisions had no recorded click time, which leaves the log time undefined, and two subjects are left with a single decision.

\autoref{tab:decisions_main} shows the regression models predicting CoT length and RT as a function of whether both posts used the same slur, as reported in Fig.~2C of the main text. The same-slur coefficient is positive and statistically significant for human subjects and for all six models: RTs were 31\% longer for same-slur pairs, and CoTs were 51--306\% longer, depending on the model. For o3 and the three open-weights models, the token models control for the log number of input tokens, which accounts for variation across pairs in how the images are represented as input. The input to Gemini 2.5 Pro and Grok 4 is a constant number of tokens. \autoref{tab:decisions_robustness} shows the robustness checks. The estimates are consistent with the main specification for all models.

\begin{table}[!htbp]
    \centering
    \footnotesize
    \caption{Identical slurs, log AI reasoning tokens, and log human RT (main specification, six models). Each token model regresses log reasoning tokens on the same-slur indicator, with input-token controls where applicable. The RT model regresses log first-click time on the same indicator. Standard errors in parentheses, clustered by subject in the RT model. *** p$<$0.001, ** p$<$0.01, * p$<$0.05.}
    \label{tab:decisions_main}
    \begin{tabular}{lccccccc}
    \toprule
    & \multicolumn{3}{c}{\textit{Closed-weights tokens}} & \multicolumn{3}{c}{\textit{Open-weights tokens}} & \\
    \cmidrule(lr){2-4} \cmidrule(lr){5-7}
     & Gemini 2.5 Pro & o3 & Grok 4 & Gemma 4 & Qwen 3.5 & Kimi-VL & Time (log) \\
    \midrule
    Same slur & 0.415*** & 0.729*** & 1.169*** & 0.882*** & 1.400*** & 1.077*** & 0.273*** \\
     & (0.005) & (0.011) & (0.016) & (0.010) & (0.016) & (0.017) & (0.012) \\
    Input tokens (log) &  & 0.113*** &  & -2.825*** & -0.099 & -1.038*** &  \\
     &  & (0.021) &  & (0.290) & (0.086) & (0.099) &  \\
    Task (log) &  &  &  &  &  &  & -0.273*** \\
     &  &  &  &  &  &  & (0.006) \\
    \midrule
    Subject FE &  &  &  &  &  &  & \checkmark \\
    \midrule
    $N$ & 27071 & 27052 & 27071 & 27071 & 27071 & 27071 & 27071 \\
    $R^2$ & 0.204 & 0.146 & 0.168 & 0.239 & 0.214 & 0.134 & 0.472 \\
    $R^2$ Adj. & 0.204 & 0.146 & 0.168 & 0.239 & 0.214 & 0.134 & 0.433 \\
    \bottomrule
    \end{tabular}
\end{table}

\begin{table}[!htbp]
    \centering
    \footnotesize
    \caption{Identical slurs, log AI reasoning tokens, and log human RT (robustness checks, six models). Panel A replicates the main specification with outliers retained. Panel B uses log last-click time as the RT outcome (outliers removed). Token-model coefficients in Panel B are identical to the main specification because the last-click measure only affects the RT outcome. Standard errors in parentheses, clustered by subject in the RT models. *** p$<$0.001, ** p$<$0.01, * p$<$0.05.}
    \label{tab:decisions_robustness}
    \begin{tabular}{lccccccc}
    \toprule
    & \multicolumn{3}{c}{\textit{Closed-weights tokens}} & \multicolumn{3}{c}{\textit{Open-weights tokens}} & \\
    \cmidrule(lr){2-4} \cmidrule(lr){5-7}
     & Gemini 2.5 Pro & o3 & Grok 4 & Gemma 4 & Qwen 3.5 & Kimi-VL & Time (log) \\
    \midrule
    \multicolumn{8}{l}{\textit{Panel A: Outliers included}} \\
    \midrule
    Same slur & 0.415*** & 0.728*** & 1.170*** & 0.878*** & 1.397*** & 1.074*** & 0.273*** \\
     & (0.005) & (0.011) & (0.016) & (0.010) & (0.016) & (0.017) & (0.013) \\
    Input tokens (log) &  & 0.116*** &  & -2.851*** & -0.074 & -1.043*** &  \\
     &  & (0.021) &  & (0.287) & (0.085) & (0.098) &  \\
    Task (log) &  &  &  &  &  &  & -0.284*** \\
     &  &  &  &  &  &  & (0.007) \\
    \midrule
    Subject FE &  &  &  &  &  &  & \checkmark \\
    \midrule
    $N$ & 27810 & 27791 & 27810 & 27810 & 27810 & 27810 & 27724 \\
    $R^2$ & 0.204 & 0.145 & 0.168 & 0.237 & 0.213 & 0.134 & 0.474 \\
    $R^2$ Adj. & 0.204 & 0.145 & 0.168 & 0.237 & 0.213 & 0.134 & 0.436 \\
    \midrule[\heavyrulewidth]
    \multicolumn{8}{l}{\textit{Panel B: Last-click outcome (outliers removed)}} \\
    \midrule
    Same slur & 0.415*** & 0.729*** & 1.169*** & 0.882*** & 1.400*** & 1.077*** & 0.301*** \\
     & (0.005) & (0.011) & (0.016) & (0.010) & (0.016) & (0.017) & (0.010) \\
    Input tokens (log) &  & 0.113*** &  & -2.825*** & -0.099 & -1.038*** &  \\
     &  & (0.021) &  & (0.290) & (0.086) & (0.099) &  \\
    Task (log) &  &  &  &  &  &  & -0.321*** \\
     &  &  &  &  &  &  & (0.005) \\
    \midrule
    Subject FE &  &  &  &  &  &  & \checkmark \\
    \midrule
    $N$ & 27071 & 27052 & 27071 & 27071 & 27071 & 27071 & 27071 \\
    $R^2$ & 0.204 & 0.146 & 0.168 & 0.239 & 0.214 & 0.134 & 0.516 \\
    $R^2$ Adj. & 0.204 & 0.146 & 0.168 & 0.239 & 0.214 & 0.134 & 0.481 \\
    \bottomrule
    \end{tabular}
\end{table}

\autoref{tab:associations_main} shows the estimates for the relationship between human RT and CoT length across all six models. Each model includes a control for the logarithm of the task number and subject fixed effects, with standard errors clustered by subject. \autoref{tab:associations_robustness} shows the robustness checks. In both panels, the estimates align with the main results in sign, significance, and magnitude.

\begin{table}[!htbp]
    \centering
    \footnotesize
    \caption{Associations between log human RT and log AI reasoning tokens (main specification, six models). Outcome is the log of human first-click time. Outliers removed. Standard errors in parentheses, clustered by subject. *** p$<$0.001, ** p$<$0.01, * p$<$0.05.}
    \label{tab:associations_main}
    \begin{tabular}{lcccccc}
    \toprule
    & \multicolumn{3}{c}{\textit{Closed-weights}} & \multicolumn{3}{c}{\textit{Open-weights}} \\
    \cmidrule(lr){2-4} \cmidrule(lr){5-7}
     & Gemini 2.5 Pro & o3 & Grok 4 & Gemma 4 & Qwen 3.5 & Kimi-VL \\
    \midrule
    Reasoning tokens (log) & 0.280*** & 0.134*** & 0.100*** & 0.183*** & 0.103*** & 0.088*** \\
     & (0.012) & (0.006) & (0.004) & (0.006) & (0.004) & (0.004) \\
    Task (log) & -0.273*** & -0.274*** & -0.274*** & -0.274*** & -0.274*** & -0.273*** \\
     & (0.006) & (0.006) & (0.006) & (0.006) & (0.006) & (0.006) \\
    \midrule
    Subject FE & \checkmark & \checkmark & \checkmark & \checkmark & \checkmark & \checkmark \\
    \midrule
    $N$ & 27071 & 27052 & 27071 & 27071 & 27071 & 27071 \\
    $R^2$ & 0.470 & 0.470 & 0.473 & 0.478 & 0.476 & 0.470 \\
    $R^2$ Adj. & 0.431 & 0.431 & 0.434 & 0.440 & 0.437 & 0.432 \\
    \bottomrule
    \end{tabular}
\end{table}

\begin{table}[!htbp]
    \centering
    \footnotesize
    \caption{Associations between log human RT and log AI reasoning tokens (robustness checks, six models). Panel A shows the main specification with outliers included. Panel B shows estimates using the log of last-click time as the outcome. Standard errors in parentheses, clustered by subject. *** p$<$0.001, ** p$<$0.01, * p$<$0.05.}
    \label{tab:associations_robustness}
    \begin{tabular}{lcccccc}
    \toprule
    & \multicolumn{3}{c}{\textit{Closed-weights}} & \multicolumn{3}{c}{\textit{Open-weights}} \\
    \cmidrule(lr){2-4} \cmidrule(lr){5-7}
     & Gemini 2.5 Pro & o3 & Grok 4 & Gemma 4 & Qwen 3.5 & Kimi-VL \\
    \midrule
    \multicolumn{7}{l}{\textit{Panel A: Outliers included (log first-click outcome)}} \\
    \midrule
    Reasoning tokens (log) & 0.285*** & 0.133*** & 0.103*** & 0.185*** & 0.103*** & 0.089*** \\
     & (0.014) & (0.007) & (0.005) & (0.007) & (0.004) & (0.004) \\
    Task (log) & -0.284*** & -0.285*** & -0.284*** & -0.285*** & -0.284*** & -0.284*** \\
     & (0.007) & (0.007) & (0.007) & (0.007) & (0.007) & (0.007) \\
    \midrule
    Subject FE & \checkmark & \checkmark & \checkmark & \checkmark & \checkmark & \checkmark \\
    \midrule
    $N$ & 27724 & 27705 & 27724 & 27724 & 27724 & 27724 \\
    $R^2$ & 0.473 & 0.473 & 0.476 & 0.479 & 0.477 & 0.473 \\
    $R^2$ Adj. & 0.435 & 0.435 & 0.438 & 0.442 & 0.439 & 0.436 \\
    \midrule[\heavyrulewidth]
    \multicolumn{7}{l}{\textit{Panel B: Last-click outcome (outliers removed)}} \\
    \midrule
    Reasoning tokens (log) & 0.324*** & 0.160*** & 0.116*** & 0.204*** & 0.120*** & 0.102*** \\
     & (0.011) & (0.005) & (0.003) & (0.006) & (0.003) & (0.003) \\
    Task (log) & -0.321*** & -0.322*** & -0.322*** & -0.322*** & -0.322*** & -0.321*** \\
     & (0.005) & (0.005) & (0.005) & (0.005) & (0.005) & (0.005) \\
    \midrule
    Subject FE & \checkmark & \checkmark & \checkmark & \checkmark & \checkmark & \checkmark \\
    \midrule
    $N$ & 27071 & 27052 & 27071 & 27071 & 27071 & 27071 \\
    $R^2$ & 0.516 & 0.517 & 0.521 & 0.527 & 0.525 & 0.516 \\
    $R^2$ Adj. & 0.480 & 0.482 & 0.485 & 0.492 & 0.491 & 0.481 \\
    \bottomrule
    \end{tabular}
\end{table}

\autoref{tab:quintiles_main} shows the regression results depicted in Fig.~4A of the main text. Each coefficient compares a quintile with the first (fewest tokens). \autoref{tab:quintiles_robustness} shows the robustness checks. The estimates rise across the quintiles for every model in both panels, as in the main specification.

\begin{table}[!htbp]
    \addedtable
    \centering
    \footnotesize
    \caption{Log human RT across quintiles of AI reasoning tokens (main specification, six models). Outcome is the log of human first-click time. Each coefficient compares a quintile of pairs, ranked by the model's reasoning tokens, with the first quintile (fewest tokens). Outliers removed. Standard errors in parentheses, clustered by subject. *** p$<$0.001, ** p$<$0.01, * p$<$0.05.}
    \label{tab:quintiles_main}
    \begin{tabular}{lcccccc}
    \toprule
    & \multicolumn{3}{c}{\textit{Closed-weights}} & \multicolumn{3}{c}{\textit{Open-weights}} \\
    \cmidrule(lr){2-4} \cmidrule(lr){5-7}
     & Gemini 2.5 Pro & o3 & Grok 4 & Gemma 4 & Qwen 3.5 & Kimi-VL \\
    \midrule
    Quintile 2 & 0.020 & 0.037** & 0.106*** & 0.058*** & 0.058*** & 0.023* \\
     & (0.011) & (0.012) & (0.012) & (0.011) & (0.011) & (0.011) \\
    Quintile 3 & 0.065*** & 0.104*** & 0.157*** & 0.141*** & 0.148*** & 0.047*** \\
     & (0.012) & (0.013) & (0.012) & (0.012) & (0.011) & (0.012) \\
    Quintile 4 & 0.131*** & 0.178*** & 0.235*** & 0.215*** & 0.235*** & 0.182*** \\
     & (0.012) & (0.013) & (0.012) & (0.012) & (0.012) & (0.012) \\
    Quintile 5 & 0.262*** & 0.249*** & 0.301*** & 0.329*** & 0.298*** & 0.230*** \\
     & (0.012) & (0.012) & (0.012) & (0.011) & (0.012) & (0.012) \\
    Task (log) & -0.273*** & -0.274*** & -0.274*** & -0.274*** & -0.274*** & -0.273*** \\
     & (0.006) & (0.006) & (0.006) & (0.006) & (0.006) & (0.006) \\
    \midrule
    Subject FE & \checkmark & \checkmark & \checkmark & \checkmark & \checkmark & \checkmark \\
    \midrule
    $N$ & 27071 & 27052 & 27071 & 27071 & 27071 & 27071 \\
    $R^2$ & 0.471 & 0.470 & 0.474 & 0.478 & 0.476 & 0.471 \\
    $R^2$ Adj. & 0.433 & 0.431 & 0.436 & 0.440 & 0.438 & 0.432 \\
    \bottomrule
    \end{tabular}
\end{table}

\begin{table}[!htbp]
    \addedtable
    \centering
    \footnotesize
    \caption{Log human RT across quintiles of AI reasoning tokens (robustness checks, six models). Panel A shows the main specification with outliers included. Panel B shows estimates using the log of last-click time as the outcome. Each coefficient compares a quintile with the first (fewest tokens). Standard errors in parentheses, clustered by subject. *** p$<$0.001, ** p$<$0.01, * p$<$0.05.}
    \label{tab:quintiles_robustness}
    \begin{tabular}{lcccccc}
    \toprule
    & \multicolumn{3}{c}{\textit{Closed-weights}} & \multicolumn{3}{c}{\textit{Open-weights}} \\
    \cmidrule(lr){2-4} \cmidrule(lr){5-7}
     & Gemini 2.5 Pro & o3 & Grok 4 & Gemma 4 & Qwen 3.5 & Kimi-VL \\
    \midrule
    \multicolumn{7}{l}{\textit{Panel A: Outliers included (log first-click outcome)}} \\
    \midrule
    Quintile 2 & 0.033** & 0.039** & 0.111*** & 0.058*** & 0.062*** & 0.028* \\
     & (0.013) & (0.014) & (0.013) & (0.013) & (0.013) & (0.013) \\
    Quintile 3 & 0.065*** & 0.117*** & 0.161*** & 0.149*** & 0.153*** & 0.038** \\
     & (0.014) & (0.014) & (0.013) & (0.013) & (0.013) & (0.013) \\
    Quintile 4 & 0.141*** & 0.187*** & 0.244*** & 0.215*** & 0.245*** & 0.179*** \\
     & (0.013) & (0.014) & (0.013) & (0.014) & (0.014) & (0.013) \\
    Quintile 5 & 0.268*** & 0.252*** & 0.310*** & 0.333*** & 0.297*** & 0.236*** \\
     & (0.014) & (0.014) & (0.014) & (0.013) & (0.014) & (0.014) \\
    Task (log) & -0.284*** & -0.284*** & -0.285*** & -0.285*** & -0.285*** & -0.284*** \\
     & (0.007) & (0.007) & (0.007) & (0.007) & (0.007) & (0.007) \\
    \midrule
    Subject FE & \checkmark & \checkmark & \checkmark & \checkmark & \checkmark & \checkmark \\
    \midrule
    $N$ & 27724 & 27705 & 27724 & 27724 & 27724 & 27724 \\
    $R^2$ & 0.474 & 0.473 & 0.477 & 0.479 & 0.477 & 0.473 \\
    $R^2$ Adj. & 0.436 & 0.435 & 0.439 & 0.442 & 0.440 & 0.436 \\
    \midrule[\heavyrulewidth]
    \multicolumn{7}{l}{\textit{Panel B: Last-click outcome (outliers removed)}} \\
    \midrule
    Quintile 2 & 0.027** & 0.058*** & 0.116*** & 0.071*** & 0.068*** & 0.028** \\
     & (0.010) & (0.011) & (0.010) & (0.010) & (0.010) & (0.010) \\
    Quintile 3 & 0.092*** & 0.132*** & 0.177*** & 0.174*** & 0.166*** & 0.063*** \\
     & (0.010) & (0.011) & (0.010) & (0.010) & (0.010) & (0.010) \\
    Quintile 4 & 0.170*** & 0.225*** & 0.271*** & 0.254*** & 0.275*** & 0.218*** \\
     & (0.010) & (0.011) & (0.011) & (0.010) & (0.011) & (0.011) \\
    Quintile 5 & 0.309*** & 0.297*** & 0.340*** & 0.367*** & 0.346*** & 0.265*** \\
     & (0.011) & (0.011) & (0.011) & (0.010) & (0.011) & (0.011) \\
    Task (log) & -0.322*** & -0.322*** & -0.322*** & -0.322*** & -0.322*** & -0.321*** \\
     & (0.005) & (0.005) & (0.005) & (0.005) & (0.005) & (0.005) \\
    \midrule
    Subject FE & \checkmark & \checkmark & \checkmark & \checkmark & \checkmark & \checkmark \\
    \midrule
    $N$ & 27071 & 27052 & 27071 & 27071 & 27071 & 27071 \\
    $R^2$ & 0.519 & 0.517 & 0.522 & 0.527 & 0.526 & 0.516 \\
    $R^2$ Adj. & 0.483 & 0.481 & 0.486 & 0.492 & 0.491 & 0.481 \\
    \bottomrule
    \end{tabular}
\end{table}

\subsection*{Pair-level regression analysis}

The main analyses use a binary indicator for whether both posts in a pair contain the same slur. To examine how specific slur pairings affect CoT length and RT, we used a categorical variable representing each unique unordered pair of slur categories (28 total). The reference category is None $\times$ Racism, which represents a pairing between a post without a slur and one containing a racial slur---the pair that elicited the fastest human decisions and among the shortest CoTs. All coefficients are therefore interpretable as the log change relative to this baseline.

\begin{figure}[!htbp]
\centering
\includegraphics[width=\textwidth]{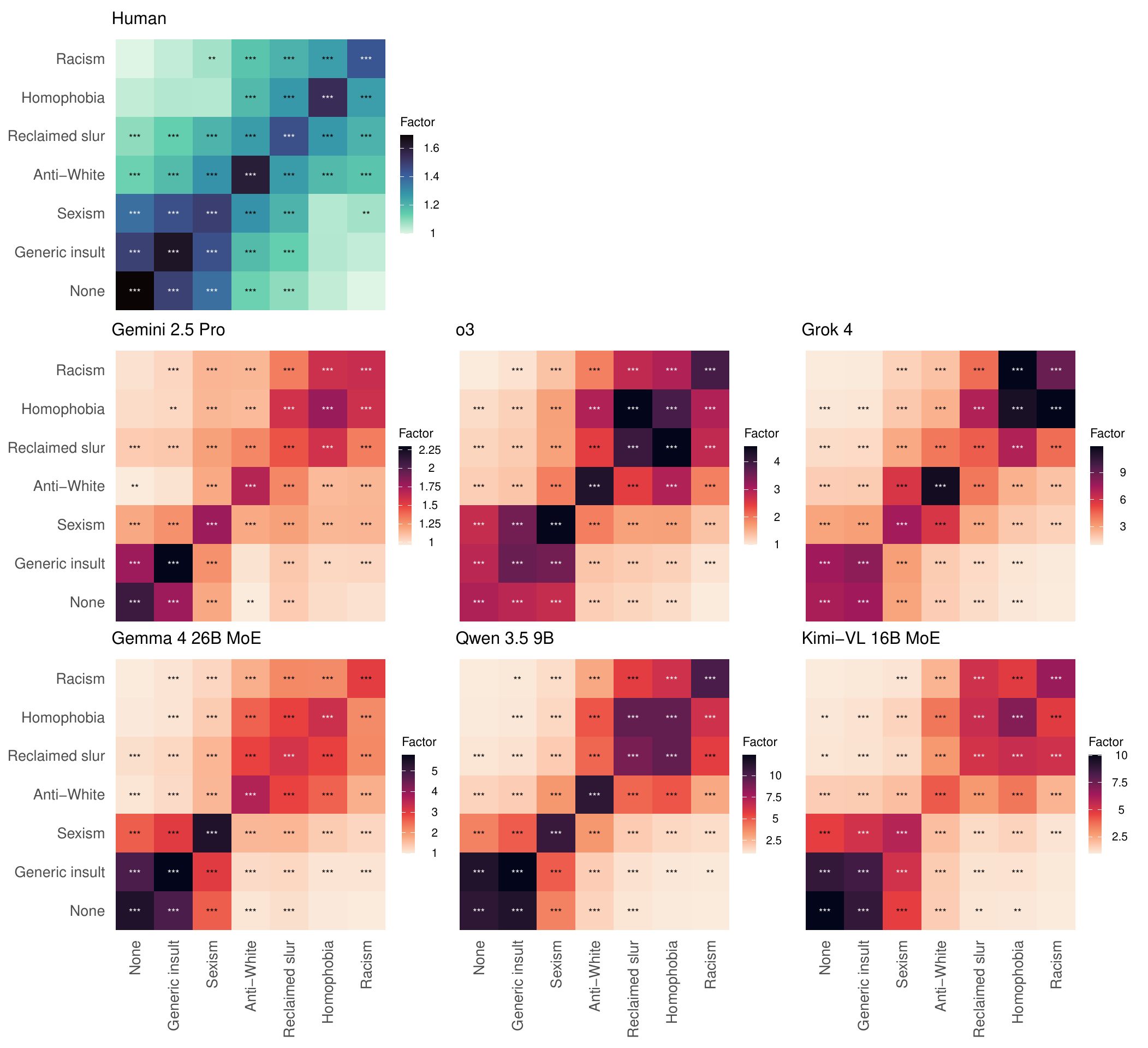}
\caption{Pair-level regression coefficients (exponentiated) for RT (top left) and CoT length (closed-weights models in the middle row, open-weights models in the bottom row). Each panel shows a 7$\times$7 symmetric matrix of slur-category pairings. Cell shading represents the multiplicative factor relative to the None $\times$ Racism reference pair (exp($\beta$)), from the regressions in \autoref{tab:decisions_pairlevel}. Stars indicate significance levels: * p$<$0.05, ** p$<$0.01, *** p$<$0.001.}
\label{fig:pairlevel_heatmap}
\end{figure}

\autoref{fig:pairlevel_heatmap} presents exponentiated regression coefficients for humans and all six models in the same 7$\times$7 symmetric heatmap format as Fig.~2A of the main text, where the color of each cell now represents the multiplicative factor by which CoT length (or RT) exceeds the None $\times$ Racism reference. A value of four, for instance, indicates that the pair elicited four times as many reasoning tokens as the reference pair.

\autoref{tab:decisions_pairlevel} reports the full regression estimates on the log scale. The token models regress log reasoning tokens on the 27 pair dummies (with input-token controls where applicable). The RT model regresses log RT on the pair dummies with a control for the log task number and subject fixed effects.

{\footnotesize
\setlength{\LTcapwidth}{\textwidth}
\setlength{\tabcolsep}{4pt}
\begin{longtable}{lccccccc}
    \caption{Pair-level slur regressions on the log scale. Each token column regresses log reasoning tokens on 27 slur-pair dummies (reference: None $\times$ Racism), with input-token controls where applicable. The RT column regresses log first-click time on the pair dummies with a control for the log task number and subject fixed effects. ``Generic'' = Generic insult; ``Reclaimed'' = Reclaimed slur. Standard errors in parentheses, clustered by subject in the RT model. *** p$<$0.001, ** p$<$0.01, * p$<$0.05.}
    \label{tab:decisions_pairlevel}\\
    \toprule
    & \multicolumn{3}{c}{\textit{Closed-weights}} & \multicolumn{3}{c}{\textit{Open-weights}} & \\
    \cmidrule(lr){2-4} \cmidrule(lr){5-7}
    & Gemini 2.5 Pro & o3 & Grok 4 & Gemma 4 & Qwen 3.5 & Kimi-VL & Time (log) \\
    \midrule
    \endfirsthead
    \multicolumn{8}{@{}l}{\textit{(continued)}}\\
    \toprule
    & Gemini 2.5 Pro & o3 & Grok 4 & Gemma 4 & Qwen 3.5 & Kimi-VL & Time (log) \\
    \midrule
    \endhead

    \endfoot
    \bottomrule
    \endlastfoot
    None $\times$ None & 0.736*** & 1.098*** & 1.983*** & 1.693*** & 2.406*** & 2.303*** & 0.525*** \\
     & (0.012) & (0.024) & (0.033) & (0.019) & (0.032) & (0.038) & (0.032) \\
    None $\times$ Generic & 0.573*** & 1.059*** & 2.022*** & 1.577*** & 2.445*** & 2.177*** & 0.399*** \\
     & (0.010) & (0.020) & (0.027) & (0.015) & (0.026) & (0.031) & (0.026) \\
    None $\times$ Sexism & 0.176*** & 1.018*** & 1.078*** & 0.969*** & 1.388*** & 1.540*** & 0.316*** \\
     & (0.010) & (0.020) & (0.028) & (0.015) & (0.027) & (0.032) & (0.026) \\
    None $\times$ Anti-White & -0.033** & 0.228*** & 0.599*** & 0.074*** & 0.524*** & 0.518*** & 0.119*** \\
     & (0.010) & (0.020) & (0.027) & (0.015) & (0.026) & (0.031) & (0.025) \\
    None $\times$ Reclaimed & 0.073*** & 0.177*** & 0.350*** & 0.131*** & 0.163*** & 0.103** & 0.089*** \\
     & (0.010) & (0.020) & (0.027) & (0.015) & (0.026) & (0.031) & (0.026) \\
    None $\times$ Homophobia & 0.019 & 0.129*** & 0.100*** & 0.021 & 0.031 & 0.094** & 0.032 \\
     & (0.010) & (0.020) & (0.027) & (0.015) & (0.026) & (0.032) & (0.024) \\
    Generic $\times$ Generic & 0.829*** & 1.279*** & 2.086*** & 1.753*** & 2.516*** & 2.137*** & 0.492*** \\
     & (0.012) & (0.024) & (0.033) & (0.018) & (0.032) & (0.038) & (0.031) \\
    Generic $\times$ Sexism & 0.237*** & 1.258*** & 1.124*** & 1.101*** & 1.600*** & 1.658*** & 0.372*** \\
     & (0.010) & (0.020) & (0.027) & (0.015) & (0.026) & (0.032) & (0.025) \\
    Generic $\times$ Anti-White & -0.006 & 0.296*** & 0.618*** & 0.172*** & 0.622*** & 0.535*** & 0.169*** \\
     & (0.010) & (0.020) & (0.027) & (0.015) & (0.026) & (0.031) & (0.025) \\
    Generic $\times$ Reclaimed & 0.083*** & 0.248*** & 0.381*** & 0.205*** & 0.241*** & 0.160*** & 0.126*** \\
     & (0.010) & (0.020) & (0.027) & (0.015) & (0.026) & (0.032) & (0.025) \\
    Generic $\times$ Homophobia & 0.033** & 0.207*** & 0.136*** & 0.084*** & 0.117*** & 0.184*** & 0.049 \\
     & (0.010) & (0.020) & (0.027) & (0.015) & (0.026) & (0.031) & (0.025) \\
    Generic $\times$ Racism & 0.036*** & 0.074*** & 0.027 & 0.084*** & 0.075** & 0.060 & 0.032 \\
     & (0.010) & (0.020) & (0.027) & (0.015) & (0.026) & (0.031) & (0.025) \\
    Sexism $\times$ Sexism & 0.575*** & 1.512*** & 2.011*** & 1.694*** & 2.375*** & 1.778*** & 0.403*** \\
     & (0.013) & (0.025) & (0.034) & (0.019) & (0.033) & (0.039) & (0.036) \\
    Sexism $\times$ Anti-White & 0.171*** & 0.672*** & 1.773*** & 0.485*** & 1.231*** & 0.702*** & 0.253*** \\
     & (0.010) & (0.020) & (0.027) & (0.015) & (0.026) & (0.031) & (0.025) \\
    Sexism $\times$ Reclaimed & 0.195*** & 0.507*** & 1.026*** & 0.493*** & 0.656*** & 0.316*** & 0.187*** \\
     & (0.010) & (0.020) & (0.027) & (0.015) & (0.027) & (0.032) & (0.025) \\
    Sexism $\times$ Homophobia & 0.134*** & 0.508*** & 0.671*** & 0.321*** & 0.433*** & 0.433*** & 0.047 \\
     & (0.010) & (0.020) & (0.028) & (0.015) & (0.027) & (0.032) & (0.025) \\
    Sexism $\times$ Racism & 0.139*** & 0.309*** & 0.521*** & 0.224*** & 0.341*** & 0.141*** & 0.067** \\
     & (0.010) & (0.020) & (0.027) & (0.015) & (0.026) & (0.031) & (0.025) \\
    Anti-White $\times$ Anti-White & 0.523*** & 1.451*** & 2.441*** & 1.293*** & 2.404*** & 1.429*** & 0.473*** \\
     & (0.012) & (0.024) & (0.033) & (0.018) & (0.032) & (0.038) & (0.029) \\
    Anti-White $\times$ Reclaimed & 0.262*** & 0.899*** & 1.413*** & 1.062*** & 1.550*** & 1.052*** & 0.233*** \\
     & (0.010) & (0.020) & (0.027) & (0.015) & (0.026) & (0.031) & (0.025) \\
    Anti-White $\times$ Homophobia & 0.124*** & 1.091*** & 0.960*** & 0.953*** & 1.633*** & 1.280*** & 0.171*** \\
     & (0.010) & (0.020) & (0.027) & (0.015) & (0.026) & (0.031) & (0.026) \\
    Anti-White $\times$ Racism & 0.134*** & 0.663*** & 0.758*** & 0.545*** & 1.061*** & 0.818*** & 0.148*** \\
     & (0.010) & (0.020) & (0.027) & (0.015) & (0.026) & (0.031) & (0.025) \\
    Reclaimed $\times$ Reclaimed & 0.373*** & 1.387*** & 1.558*** & 1.162*** & 2.186*** & 1.643*** & 0.372*** \\
     & (0.012) & (0.024) & (0.033) & (0.018) & (0.032) & (0.038) & (0.032) \\
    Reclaimed $\times$ Homophobia & 0.465*** & 1.514*** & 1.956*** & 1.066*** & 2.256*** & 1.698*** & 0.241*** \\
     & (0.010) & (0.020) & (0.027) & (0.015) & (0.026) & (0.031) & (0.026) \\
    Reclaimed $\times$ Racism & 0.287*** & 1.037*** & 1.481*** & 0.779*** & 1.744*** & 1.664*** & 0.188*** \\
     & (0.010) & (0.020) & (0.028) & (0.015) & (0.027) & (0.032) & (0.027) \\
    Homophobia $\times$ Homophobia & 0.580*** & 1.355*** & 2.425*** & 1.198*** & 2.254*** & 1.947*** & 0.436*** \\
     & (0.013) & (0.025) & (0.034) & (0.019) & (0.033) & (0.039) & (0.034) \\
    Homophobia $\times$ Racism & 0.475*** & 1.093*** & 2.474*** & 0.765*** & 1.856*** & 1.564*** & 0.230*** \\
     & (0.010) & (0.020) & (0.027) & (0.015) & (0.026) & (0.032) & (0.026) \\
    Racism $\times$ Racism & 0.484*** & 1.362*** & 2.197*** & 1.089*** & 2.323*** & 1.880*** & 0.359*** \\
     & (0.012) & (0.024) & (0.033) & (0.019) & (0.032) & (0.038) & (0.034) \\
    Input tokens (log) &  & 0.140*** &  & -3.333*** & 0.025 & -0.908*** &  \\
     &  & (0.016) &  & (0.186) & (0.056) & (0.075) &  \\
    Task (log) &  &  &  &  &  &  & -0.274*** \\
     &  &  &  &  &  &  & (0.006) \\
    \midrule
    Subject FE &  &  &  &  &  &  & \checkmark \\
    \midrule
    $N$ & 27071 & 27052 & 27071 & 27071 & 27071 & 27071 & 27071 \\
    $R^2$ & 0.454 & 0.522 & 0.595 & 0.688 & 0.665 & 0.505 & 0.488 \\
    $R^2$ Adj. & 0.453 & 0.522 & 0.595 & 0.688 & 0.664 & 0.504 & 0.450 \\
\end{longtable}
}

\subsection*{Agreement with human choices}

\begin{figure}[!htbp]
\centering
\includegraphics[width=\textwidth]{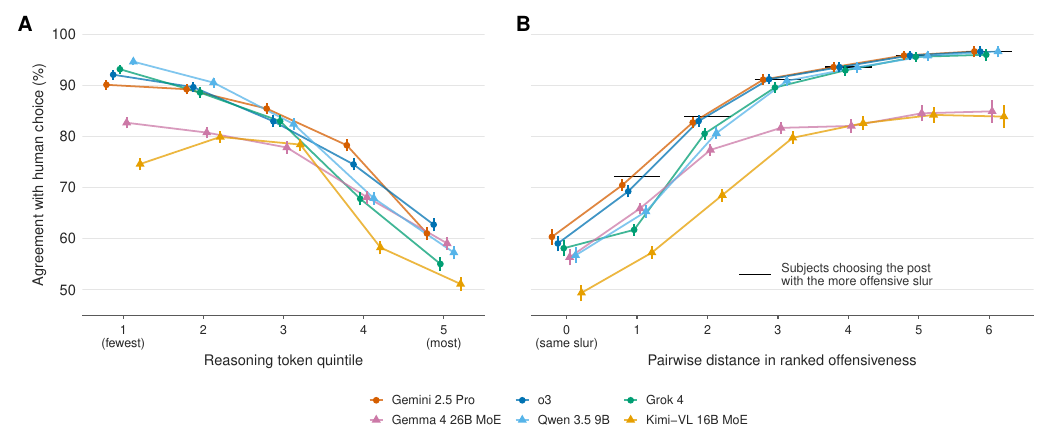}
\caption{Agreement with human choices. Each point shows the share of pairs on which a model chose the same post as the subject who judged the pair. (A) By quintile of each model's reasoning tokens, as in Fig.~4A of the main text. (B) By the distance between the two slurs in the offensiveness ranking, from 0, when both posts contain the same term, to 6, for None $\times$ Racism, as in Fig.~3 of the main text. Black lines show the share of subjects who chose the post with the more offensive slur, which is the agreement a model would reach by always choosing that post. Error bars represent 95\% confidence intervals, with standard errors clustered by subject. $N = 27{,}071$ pairs ($N = 27{,}052$ for o3).}
\label{fig:agreement}
\end{figure}

The task has no correct answer, so we cannot assess whether longer reasoning yielded better decisions. Each pair was judged by only one subject, so we also cannot directly measure disagreement among people about the same decision, and lower agreement does not imply that additional reasoning leads to worse decisions. While there were sometimes substantial differences in perceived offensiveness between the two posts in a pair, reasonable people could reach different conclusions about many of the examples used in the experiment. We can, however, assess how often each model chose the same post as the subject who judged each pair and whether this agreement is related to reasoning effort. Overall agreement ranged from 68.5\% (Kimi-VL 16B MoE) to 80.8\% (Gemini 2.5 Pro). It was lowest on the pairs where models reasoned most (\autoref{fig:agreement}A): 51--63\% in the top quintile of reasoning tokens, compared with 75--95\% in the bottom quintile. These were mostly difficult pairs given that LRMs reasoned most when the slurs were the same or adjacent in offensiveness. Agreement rose with the distance between the two slurs (\autoref{fig:agreement}B), from 49--60\% when both posts used the same slur to 84--97\% when the slurs were furthest apart. At distances of three or more, Gemini 2.5 Pro, o3, Grok 4, and Qwen 3.5 9B chose the post with the more offensive slur in at least 98\% of pairs, so their agreement was within two percentage points of the share of subjects who made the same choice (see black lines). This is consistent with related findings that LLMs can place weight on the presence of slurs similar to that of human subjects \citep{davidson-multimodal-2025}. Gemma 4 26B MoE and Kimi-VL 16B MoE leveled off lower because they appeared to favor the first post, choosing it in 61\% and 66\% of pairs, compared with 49\% for human subjects. Since this pattern was present even when the slurs were far apart, it suggests that these models were sensitive to the order of the posts, an irrelevant factor, and not only on the most difficult pairs. Overall, this analysis shows that most models tended to make the same choice as subjects when offensiveness diverged, whereas agreement was lower on the difficult pairs. This finding is consistent with greater ambiguity on these pairs, as judgments entailed the interpretation of secondary factors that might not be directly associated with perceived offensiveness.

\section*{Reasoning traces}

\textbf{\textit{Warning: This section includes offensive language.}}

\label{sec:reasoning_examples}

The following examples are summaries of the CoT from Gemini 2.5 Pro that were sampled for the qualitative analysis. Each summary is provided along with a brief explanation, including the factors that were coded by two or more raters during the qualitative analysis. The number of tokens used in the CoT is reported at the end of each example. Note that this is the length reported by Gemini and does not necessarily correspond to the number of words in the text summary. The summaries tend to follow a similar structure, beginning with a brief restatement of the instructions, followed by a description of the two posts, and finally an explanation for the decision about which post to prioritize. The summaries are reproduced verbatim, including the model's original Markdown-style formatting (asterisks). The first three examples feature explanations for posts with different slurs, and the next three feature posts that used the same slur.

\subsection*{Different slurs}
The first example demonstrates how slurs can be used to justify a decision when there is a clear disparity in offensiveness, although raters also coded the topic as being mentioned in the explanation.

\begin{quote}
**My Content Moderation Analysis** Okay, here's my breakdown of this image moderation task. I need to act as a content moderation expert and decide which image, A or B, requires the *highest* priority for manual review. The policy is clear: it targets abusive behavior based on protected characteristics like race, ethnicity, and so on. First, I examined Image A. It's a tweet about pre-fight trash talk. The author thinks the fighters are dramatic, like they should be in drama school, and it is standard sports commentary about behavior. It doesn't seem to violate the policy at all. Then, I focused on Image B. This is a far more serious case. It's about someone cutting in line at a coffee shop, but the language used is abhorrent. The user uses the N-word and associates that word with someone's entitled behavior. That's a clear, direct, and severe violation of the hate speech policy, as it's targeted directly at the person's race. Comparing the two, Image A is likely a low-priority, perhaps not even a violation. Image B, however, is a clear and egregious violation of the policy. The use of a racial slur demands immediate attention. Therefore, Image B needs to be prioritized for manual review. The output needs to be a single character, and based on this analysis, the answer is undoubtedly **B**. [CoT length: 637 tokens]
\end{quote}

The next example shows another case where the two posts use different slurs. In this case, one post uses an anti-White term (so-called ``reverse racism'') and the other an anti-Black slur. The slur is the only factor coded as relevant by the raters.

\begin{quote}
**Content Moderation Prioritization: A Thought Process** Alright, let's break this down. My task is clear: I'm acting as a content moderator, and I need to prioritize one image over another for manual review based on a policy focused on preventing abuse and targeting of protected groups. The policy is centered around abusive behavior, particularly slurs, targeting individuals or groups based on characteristics like race. Got it. First, I examined Image A. The tweet uses "cracker," a racial slur directed at white people. The policy explicitly prohibits this kind of targeting, so I flagged it immediately. It's a violation, no question. The "cracker" statement is a slur and thus problematic. Next, I turned my attention to Image B. This one hits harder. "Nigger" is used in reference to Black individuals. The phrase also links "nigger" to "entitled behavior". This is a historically loaded term with a brutal history of oppression, directly targeting a historically marginalized group. This violates the policy in a more severe way. Comparing the two, both are violations. But the n-word is something else entirely. It carries so much historical baggage and represents a level of hate I have to consider more serious. While "cracker" is also a slur, it doesn't carry the same weight as the n-word. Therefore, because the policy emphasizes protecting historically marginalized groups, the content in Image B is a higher priority. It targets Black people with a severe slur. Thus, I'll prioritize Image B for review, because it presents a more severe violation of the policy. The final output is simple: B. [CoT length: 749 tokens]
\end{quote}

Another example compares a post using a sexist insult to one without any slurs. It was coded as mentioning the slur and the topic.

\begin{quote}
**Content Moderation Analysis: Prioritizing Image for Manual Review** Okay, here's the situation. I've got two images (A and B) and need to figure out which one warrants immediate manual review, based on a specific content moderation policy. The policy's pretty clear: it's all about prohibiting abuse targeting individuals or groups based on protected categories – race, ethnicity, sexual orientation, you get the idea. I need to output a single letter: 'A' or 'B'. First, let's break down Image A. It's a tweet from "Seth" that says, "Finally, some common sense climate policies. Eco-warrior bitches always want a handout." Alright, right off the bat, I see a red flag. The policy specifically calls out "abuse" and the term "bitches" is definitely a derogatory slur, especially when used to target a group. I'd argue this is an attack on a group – specifically framed as women – through a gendered slur. The policy explicitly prohibits abuse based on *gender*, so this immediately triggers my concern. Now, let's look at Image B. This one's from "Jayvon" and reads, "This pre-fight trash talk is crazy! Some fighters should just switch to drama school." Hmm, it's critical, but not abusive in the way the policy defines it. The target is "some fighters" and the commentary is about their behavior. "Fighters" aren't a protected category, and it doesn't seem to be targeting anyone based on race, gender, religion, etc. It's really just run-of-the-mill sports commentary, albeit a bit snarky. Comparing the two, it's a no-brainer. Image A is a clear potential violation due to the use of a gendered slur to attack a group. Image B, on the other hand, is likely fine. Therefore, I need to prioritize Image A. The output needs to be a single character, so the answer is **A**. [CoT length: 714 tokens]
\end{quote}

\subsection*{Identical slurs}

In this example, both posts use the same racial slur. The model mentions the race of the users as a key factor in the adjudication. The summary was coded as mentioning four factors: slur, user gender, user race, and topic.

\begin{quote}
**Prioritization of Content for Manual Review** Okay, here's the breakdown of my thinking. My role is to make the call on which image needs urgent manual review, based on the platform's content moderation policy: no targeting individuals or groups with abuse based on perceived membership in a protected category. We're looking for abuse *motivated* by hatred, prejudice, or intolerance. First, I looked at Image A. It's a tweet from someone named "Sarah" (profile pic of a white woman), and the text is a harsh racial slur targeting "studio execs." It's clear as day: the language is explicitly offensive, a blatant use of a racial slur coming from an out-group member. This is a severe violation, motivated by prejudice. I see a high likelihood of it being hate speech and it should be prioritized. Then I considered Image B. This one is from "Tyra" (profile pic of a Black woman), and the tweet includes the same racial slur, but the context is vastly different. It's positive commentary about a basketball play. The user is also Black. Context matters *immensely*. While the word is undeniably offensive, the intent here *seems* to be more cultural, perhaps slang within a specific community. It's not the same as the overt, targeted attack in Image A. It might even be reclaimed. This isn't the hate-motivated behavior our policy is really aimed at. Comparing the two, the severity is clear. Image A is a clear-cut case of hate speech, whereas Image B is a much more nuanced situation. The policy focuses on abuse motivated by hatred, prejudice, or intolerance. In that context, Image A is the *clearer* violation and more likely to be the type of abuse the policy covers. Therefore, Image A should absolutely be prioritized for manual review. **Final decision is A.** [CoT length: 959 tokens]
\end{quote}

The following reasoning trace compares two posts using the same racial slur. Here, interactions with the posts are mentioned in the explanation. The trace was coded as mentioning slurs, topic, engagement, and replies.

\begin{quote}
**My Assessment of Potentially Abusive Content** Okay, here's how I'm thinking about this. I've got two images to review, and the goal is to prioritize which one needs a manual review based on our content moderation policies. The policy is pretty clear: it's about protecting individuals and groups from targeted abuse based on protected characteristics like race, ethnicity, sexual orientation, etc. The policy specifically mentions "direct attack," which is key here. Let's break down each image: **Image A:** The text reads, *"That fucking offsides call was pure robbery! Can these nigger refs even see straight?"* My first thought is, yikes. This *clearly* uses a racial slur, "nigger," to directly attack a group of people, the referees. It's a blatant violation of the policy, a direct attack based on the perceived race of the refs. No question about it. **Image B:** The text reads, *"Saw that movie everyone won't shut up about. A waste of 2 hours. Who funds these nigger directors?"* Again, a racial slur, "nigger," is used. This time, it's aimed at the directors of a movie. Same as Image A, it's a direct attack based on race. Equally as problematic. This is also a clear violation. Now, we're asked to prioritize. Both are bad. Both violate the policy severely. They both use the same highly offensive racial slur to target groups based on their race. So, the core violation - the use of the slur and the intent to attack a group - is identical. The key is to look for subtle differentiators. Engagement is critical here. Image A has 2 likes. Image B has 26 likes *and* a reply in agreement. That level of engagement is a red flag. Higher engagement equals higher potential harm. More people are seeing it, and it's being amplified by the reply. It's reached a wider audience, and the message is being legitimized, to some extent, by that reply. The context of the posts, frustrations with a sports call and a movie, don't really affect the severity of the *racial abuse* itself. The target is a group in both. Neither specifically targets a named individual. Given the higher engagement on Image B, it has a more significant potential for broader impact and causing harm. I'd prioritize reviewing that one. Both are severe violations, but if I have to choose one to prioritize, the one with more engagement is the logical choice, due to the increased probability of harm. **Final decision: Image B gets the higher priority for manual review.** [CoT length: 701 tokens]
\end{quote}

Some of the longest CoTs occurred when the model was forced to decide between two posts that did not contain any extremely hateful material, such as the following explanation for a decision about two posts using a generic insult. Here, raters coded slur (raters were instructed to select this category when the generic insult was mentioned), cursing, topic, and engagement as relevant to the decision. Note how the model also alludes to connotations of the names, but the summary was not coded as mentioning the user race or gender.

\begin{quote}
**Here's how I'm thinking about this:** Okay, so I've got to determine which image, A or B, is *more* likely to violate the policy. It's all about behavior targeting individuals or groups with abuse, specifically based on protected categories. I'm focusing on the *reason* for the abuse. First, I looked at Image A. It's a tweet using profanity about baseball players. The problem is the word "asshole," and calling out the "asshole batters". But, there's nothing in there that's targeting anyone based on race, gender, religion, or anything like that. It's a general complaint about performance. It's just an angry fan using some harsh language. So, Image A seems pretty harmless from a policy perspective, not a violation. It is, therefore, low priority. Then I moved to Image B. Again, it has some profanity ("fucking", "assholes") and is a complaint about management related to basketball. Again, no specific targets based on protected categories. So, it is similar. Not a violation either. Low priority, again. Now, the trick: I must choose one to be the *higher* priority. Both seem equally low-priority violations of the policy. I have to find a difference. I considered the names ("Jamal" and "Scott"). Are there potential biases based on those names? But the text itself isn't *about* race or ethnicity. So, that's not it. I then had to think if it's about the kind of targets. One is about "batters" (players) and the other is about "management" (an organizational body). Is it more problematic to insult individuals than groups? Could be a slight point of difference but in this case, "management" is still a group of people and in the context given is no more or less concerning than "batters." Then there is the issue of reach and engagement: So, between the two, image B has more engagement, a factor I think is relevant to a real moderation situation. A post validated by others is more likely to be more widely seen and more likely to be taken in the wrong context by users. So, although it's very close, I'm choosing 'B'. This feels most consistent with standard moderation principles - prioritizing content with greater potential for impact, even if it's unlikely to be a policy violation in the provided context. [CoT length: 2182 tokens]
\end{quote}

\end{document}